\pdfoutput=1
\documentclass[11pt]{article}

\usepackage{ACL2023}

\usepackage{times}
\usepackage{latexsym}

\usepackage[T1]{fontenc}
\usepackage[utf8]{inputenc}

\usepackage{microtype}

\usepackage{inconsolata}

\usepackage{booktabs}
\usepackage{tabularx}
\usepackage{multicol}
\usepackage{multirow}
\usepackage{graphicx}
\usepackage{subfig}
\usepackage{hyperref}
\usepackage{float}

\usepackage{xcolor,colortbl}
\definecolor{darkgreen}{HTML}{008A22}
\definecolor{green}{HTML}{76d275}
\definecolor{lightgreen}{HTML}{b2fab4}
\definecolor{lightestgreen}{HTML}{d7ffd9}

\usepackage{color,soul}
\usepackage{mathtools}
\title{Boosting Distress Support Dialogue Responses with Motivational Interviewing Strategy}

\author{Anuradha Welivita, and Pearl Pu\\
  School of Computer and Communication Sciences \\
  École Polytechnique Fédérale de Lausanne \\
  Switzerland \\
  \texttt{\{kalpani.welivita,pearl.pu\}@epfl.ch}\\}

\begin{document}
\maketitle
\begin{abstract}

AI-driven chatbots have become an emerging solution to address psychological distress. Due to the lack of psychotherapeutic data, researchers use dialogues scraped from online peer support forums to train them. But since the responses in such platforms are not given by professionals, they contain both conforming and non-conforming responses. In this work, we attempt to recognize these conforming and non-conforming response types present in online distress-support dialogues using labels adapted from a well-established behavioral coding scheme named  Motivational Interviewing Treatment Integrity (MITI) code and show how some response types could be rephrased into a more MI adherent form that can, in turn, enable chatbot responses to be more compliant with the MI strategy. As a proof of concept, we build several rephrasers by fine-tuning Blender and GPT3 to rephrase MI non-adherent \textit{Advise without permission} responses into \textit{Advise with permission}. We show how this can be achieved with the construction of pseudo-parallel corpora avoiding costs for human labor. Through automatic and human evaluation we show that in the presence of less training data, techniques such as prompting and data augmentation can be used to produce substantially good rephrasings that reflect the intended style and preserve the content of the original text.

\end{abstract}

\section{Introduction}

\begin{figure}[t!]
  \centering
  \includegraphics[width=\linewidth]{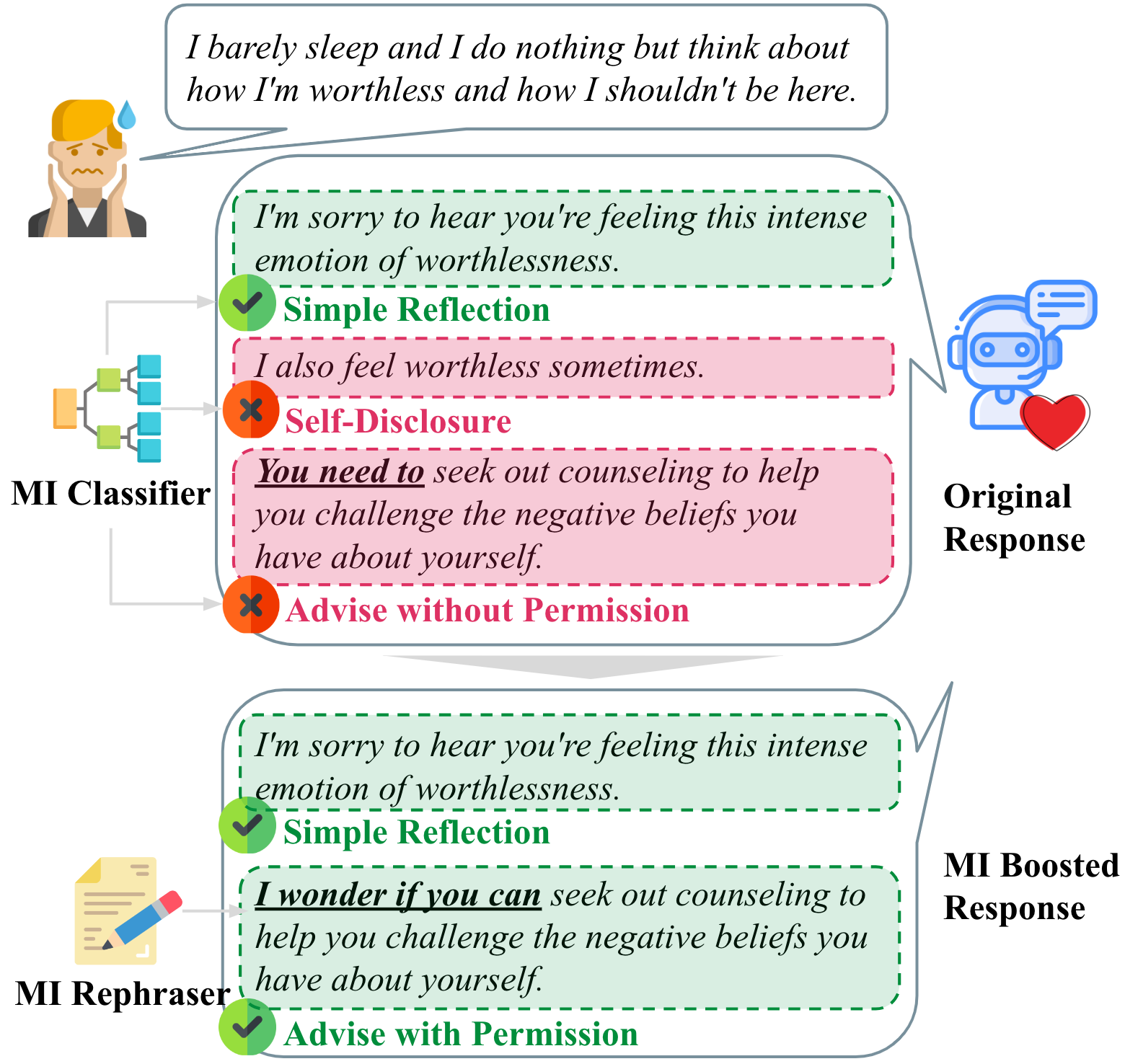}
  \caption{Example of detecting unfavourable and favourable response types in distress support dialogues and boosting the responses by omitting unfavourable responses or rephrasing them into more favourable ones.}
  \label{fig:example}
  \vspace{-0.5cm}
\end{figure}

%\begin{figure*}[t!]
%  \centering
%  \includegraphics[width=\linewidth]{Images/Schema_diagram.png}
%  \caption{The end-to-end workflow in boosting distress support dialogue responses with MI strategy.}
%  \label{fig:workflow}
%\end{figure*}

% , especially the younger generation could be
%  and social media

Demands of the modern world are increasingly responsible for causing severe psychological distress in people. World Health Organization estimates psychological distress affects 29\% of people in their lifetime \cite{mental-disorder}. The shortage of mental health workers and the stigma associated with mental health further demotivates people from actively seeking help. With the expansion of the internet, many people are seen resorting to peer support platforms such as Reddit and Talklife to vent their distress.\footnote{\url{www.reddit.com}; \url{www.talklife.com}} The anonymity associated with these platforms makes it easier for people to discuss their concerns without being affected by the stigma. Distress consolation through AI-driven chatbots has also become an emerging solution \cite{woebot, wysa, mousavi}. Due to the lack of availability of large-scale psycho-therapeutic conversations, researchers are using data scraped from online peer support forums to train such chatbots \cite{suicide, heal}. High levels of perceived empathy and information richness make them good candidates for training \cite{nambisan2011information, de2014mental, sharma2020engagement, sharma2020computational}. But since peers are not professionals, the responses contained in such forums can sometimes be unfavourable to address distress (e.g. confrontations, judgments, orders etc.). So, using this data can have severe risks. One solution for this is identifying favourable and unfavourable response types that appear in distress support dialogues and developing automatic means that can propose omission or rephrasing of such unfavourable response types. Figure \ref{fig:example} shows an example.

% Studies suggest peer support dialogues are good candidates for training due to high levels of perceived empathy and information richness 

% , which is also the focus of this work,

% such chatbots 

% both resorting to peer support forums and 

% Some studies focus on attributes such as perceived empathy and information richness in peer support dialogues that suggest they are good candidates for training such chatbots \cite{nambisan2011information, de2014mental, sharma2020engagement, sharma2020computational}. 

% causing psychological burdens and 

% And peers who have had similar experiences are seen to actively respond to such posts containing distress. 

%  for this situation

% both resorting to peer support forums and developing artificial agents based on such data bear severe risks since they can contain responses that can be unfavourable for addressing distress (e.g. confrontations, judgments, orders etc.). 

%This work focuses on developing automatic means of detection and rephrasing of such unfavourable responses.

%To attain this, it is important to identify favourable and unfavourable response types that appear in distress support dialogues. A good source where such dialogues can be curated is peer support forums such as Reddit. Peers actively engage in Reddit to support others going through distressful situations in life. Since peers are not professionals unfavourable response types are much likely to be detected in such platforms. To have a comparative understanding 

To analyze the types of responses in distress support dialogues, we use labels adapted from a well-established behavioral coding system named  Motivational Interviewing Treatment Integrity (MITI) code \cite{miti_4_2_1}. It is used in psychology to evaluate how well a mental health provider responds. Specific response types from the MITI code have shown to increase the likelihood of positive health outcomes \cite{perez2018analyzing, gaume}. It defines favourable response types such as \textit{Questioning}, \textit{Reflecting}, and \textit{Advising with permission} and unfavourable response types such as \textit{Advising without permission}, \textit{Confronting}, and \textit{Self-Disclosing (extra-session)}. In our previous work, we developed a dataset called the MI dataset, to have a comparative understanding of the differences between online support provided by peers and trained counselors. For this, we hired professional counselors to annotate responses given by peers and counselors with labels derived from the MITI code. During analysis, we observed that peers' responses tend to be more supportive, and encouraging than counselors’ (as observed by the increased percentage of \textit{Support} and \textit{Affirm} labels). But it was also observed that important therapeutic techniques, such as asking more \textit{open questions} than \textit{closed} ones, \textit{reflections}, \textit{giving information}, \textit{advices with permission}, and \textit{emphasizing speaker’s autonomy} were lacking in peers' responses and hence require further boosting. One of the major observations was that among the advises given by the peers, 92.86\% of them belonged to the category \textit{Advise without permission}, which is MI non-adherent. This percentage was lower in counselor responses, but still accounted for 77.22\% of the advises given by counselors.

In this work, we aim to detect such \textit{Advise without permission} responses among distress support dialogues and build a rephraser that can rephrase such responses into \textit{Advise with permission}, which is more MI-adherent. First, we detect such responses through a classifier trained on an augmented version of the MI dataset. Next, as we do not have human written responses rephrasing \textit{Advise without permission} responses into \textit{Advise with permission}, we use automatic methods such as template-based replacement and retrieval to construct a pseudo-parallel training corpus containing pairs of \textit{Advise without permission} and \textit{Advise with permission} sentences. Since rephrasing is a labor-intensive task compared to labeling and we require professionally trained counselors to do this in the distress consolation setting, using our already labeled dataset to construct a pseudo-parallel corpus saved us both time and cost. We apply the same methods on the augmented version of the MI dataset to form a much larger pseudo-parallel training corpus and use these corpora to fine-tune BlenderBot \cite{blender} and GPT3 \cite{gpt3}. Some of the models we fine-tune incorporate different forms of prompting with the aim of obtaining a better outcome with less training examples. We evaluate the rephrasers using automatic and human evaluation. The results mainly show when the training dataset is small, prompting improves the performance of the rephrasers across style transfer and semantic similarity dimensions. They also suggest that when the training dataset is large (in our case through data augmentation), pseudo-parallel data generated through simpler methods such as template replacement produce better results.

%can enable the models to generate substantially good rephrasings. 

% Human evaluation is conducted by recruiting professionally trained counsellors to rate the quality of the rephrasings in the style transfer and semantic similarity dimensions. The results suggest that rephrasers that utilize GPT3 perform better than those that utilize Blender. We show that when the training dataset is small, prompting improves the performance of the rephrasers across the above dimensions. We also show that when the training dataset is large (in our case through data augmentation), pseudo-parallel data generated through simpler methods such as template replacement can enable the models to generate substantially good rephrasings. 

% few-shot learning

% The pseudo-parallel dataset was further augmented to form a much larger training corpus. 

%Figure \ref{fig:workflow} denotes this end-to-end workflow in boosting distress support dialogue responses with Motivational Interviewing strategy.

%We also discuss how this methodology could be applied to boost chatbot responses by making them more compliant with the MI strategy.

Our contributions are four-fold. 1) We develop an MI classifier that can predict 15 different favourable and unfavourable response types derived from the MITI code. 2) We propose a methodology to rephrase responses detected as \textit{Advise without Permission} into more MI-adherent \textit{Advise with Permission}. We show how this can be done in the absence of human written rephrasings by developing pseudo-parallel corpora using different automatic methods. 3) We evaluate these rephrasers using automatic and human evaluation and show how prompting and data augmentation can improve the performance of the rephrasers when there is less training data. 4) Finally, we discuss how this method can be applied to boost chatbot responses, making them more compliant with the MI strategy. Our code and the datasets can be found at \url{https://github.com/anuradha1992/Boosting-with-MI-Strategy}

%similar type of methodology could be used to boost dialogue responses across the whole MI spectrum of labels.  

\section{Related Work}
\label{sec:lit_review}

%To do

%The field of TST involves traditional linguistic approaches for style transfer such as using term replacement. They require more domain-specific templates, hand-featured phrase sets that express a certain attribute (e.g., friendly), and sometimes a look-up table of expressions with the same meaning but multiple different attributes. 

%Though the field of text style transfer 

% osman2003, power2003, 
% , nikolov2019

Rephrasing responses recognized as \textit{Advise without Permission} into \textit{Advise with Perrmission} can be identified as a sub-task falling under the task of Text Style Transfer (TST), in which the goal is to automatically control the style attributes (e.g. sentiment, politeness, humor, etc.) of text while preserving the content \cite{tst_survey}. The field of TST involves traditional linguistic approaches as well as deep learning approaches. Traditional approaches to TST rely on term replacement and templates \cite{mairesse2011, sheikha2011}. With the success of deep learning, various neural methods have been recently proposed for TST. Given datasets in which there are direct mappings between the text of the source style and the text of the target style, which are referred to as parallel corpora, standard sequence-to-sequence models are often directly applied for TST \cite{GYAFC, shang2019, xu2019}. But parallel corpora are challenging to find because the development of such data often requires costly human labor. Thus, TST on non-parallel corpora has become an emerging area of research \cite{li2018, jin2019, liu2022}. 

%Various data augmentation techniques such as back-translation \cite{sennrich2016} have been proposed due to the lack of availability of parallel datasets for this task \cite{GYAFC, zhang2020}. 

% . They require more domain-specific templates, hand-featured phrase sets that express a certain attribute, and sometimes a look-up table of expressions with the same meaning but multiple different attributes \cite{osman2003, power2003, mairesse2011, sheikha2011}. amazon, 

% gender \cite{prabhumoye}, 

Parallel and nonparallel datasets have been proposed for common sub-tasks of TST such as sentiment \cite{yelp}, topic \cite{yahoo}, formality \cite{GYAFC}, politeness \cite{madaan2020}, and humor \cite{gan2017} transfer. But to the best of our knowledge, this is the first attempt at introducing a new sub-task and releasing an nonparallel corpus for style transfer between MI non-adherent \textit{Advise without Permission} and MI adherent \textit{Advise with Permission} responses. This task is more challenging than the other sub-tasks because it requires the expertise of professional counselors to generate training data. In this work, we release a nonparallel corpus that can be utilized for this task, which is annotated by professional counselors. We also show how automatic methods could be applied to create pseudo-parallel corpora using this dataset, which can be used to train neural models for this task.  

% due to the difficulty in generating data. Generating data for this task requires the expertise of professionally trained counselors. 

%  whose labor is costly to obtain

\section{Datasets}
\label{sec:datasets}

% To have a comparative understanding between the responses given by counselors and peers in support of those undergoing distress, in our previous work, we curated dialogues from two online support platforms.

% Descriptive statistics about these datasets are included in the appendices. 

For this work, we used dialogues curated from two online support platforms. The first one is CounselChat \href{https://counselchat.com/}{(counselchat.com)}, in which verified counselors respond to distress-related posts. The CounselChat dataset available publicly \footnote{\url{https://github.com/nbertagnolli/counsel-chat}} contains 2,129 post-response pairs spanning 31 distress-related topics. We also curated dialogues from a carefully selected set of 8 subreddits: \textit{mentalhealthsupport}; \textit{offmychest}; \textit{sad}; \textit{suicidewatch}; \textit{anxietyhelp}; \textit{depression}; \textit{depressed}; and \textit{depression\_help}, which are popular among Reddit users to vent their distress. This dataset, which we call RED (Reddit Emotional Distress), contains 1,275,486 dyadic conversations having on average of 2.66 turns per dialogue.

In our previous work, we recruited professional counselors to annotate a subset of 1,000 dialogues each from CounselChat and RED datasets with labels adapted from the MITI code 2.0 \cite{moyers} and 4.2.1 \cite{miti_4_2_1}. We call this the MI dataset. We used 15 labels for annotation. They are elaborated in the appendices. Out of them, we are interested in the labels \textit{Advise with Permission} and \textit{Advise without Permission}, which are respectively considered MI-adherent and MI non-adherent response types. The MI dataset contains 16,811 annotated responses, out of which 2.87\% (484) and 13.5\% (2,285) responses are labeled as \textit{Advise with Permission} and \textit{Advise without Permission}, respectively.

To further augment the MI dataset, we used automatic labeling to expand the 15 labels into unlabeled dialogue responses from CounselChat and RED datasets. We used two automatic methods for this purpose: 1) N-gram-based matching; and 2) Similarity based retrieval. 

%\subsection{N-gram Based Matching}

%During N-gram-based matching,  by analyzing the responses in the MI dataset

\textbf{N-gram Based Matching:} By tokenizing the responses in the MI dataset and computing the frequencies, we discovered the most frequent N-grams (four-grams and five-grams) occurring among the 15 labels. Examples of them are shown in the appendices. Next, we searched for the presence of these indicative N-grams (first five-gram and then four-grams) among individual sentences that appear in dialogue responses of the unlabeled CounselChat and RED datasets. If an indicative N-gram was found in a sentence, we labeled that sentence with the label that N-gram is indicative of. The sentences with overlapping labels were discarded due to ambiguity. In this way, we were able to automatically label 1,918 and 340,361 sentences in CounselChat and RED datasets, respectively. 

%We also encountered 518 and 53,196 sentences in CounselChat and RED datasets respectively that had overlapping labels, which were discarded due to ambiguity. 

% The numbers corresponding to each label are denoted in the appendices.

% The most overlaps were detected among the response pairs: \textbf{Give Information} - \textbf{Advise without Permission}; and \textbf{Advise with Permission} - \textbf{Advise without Permission}.

%\subsection{Similarity Based Retrieval}

\textbf{Similarity Based Retrieval:} For each unlabeled sentence among the responses in CounselChat and RED datasets, we computed the cosine similarity with each of the labeled sentences in the MI dataset. Next, for each unlabeled sentence, we retrieved the labeled sentences whose cosine similarity is higher than a certain threshold (the thresholds were different for each of the 15 labels, which were selected after manually inspecting randomly selected pairs of unlabeled and labeled sentences corresponding to different labels). Next, we used a majority voting scheme to select the label we can associate the unlabeled sentence with. When we encountered ties, we computed the average similarities across the clusters of retrieved sentences with different labels that held a tie and selected the label based on maximum average similarity. Using this method, we were able to automatically annotate 2,881 and 1,196,012 sentences in CounselChat and RED datasets, respectively. 

% The numbers corresponding to each label are denoted in the appendices. 

% The technical details are elaborated in the appendices. 

Using the union and the intersection of the labels retrieved from N-gram-based matching and similarity-based retrieval and combining them with the gold labels from the MI dataset, we created two augmented-labeled MI datasets having 1,378,469 and 84,052 labeled sentences, respectively. For simplicity, we will refer to them as MI Augmented (Union) and MI Augmented (Intersection) datasets. 

%Using the union and the intersection of the labels retrieved from N-gram-based matching and similarity-based retrieval and combining them with the gold labels from the MI dataset, we created two augmented-labeled MI datasets having 1,378,469 and 84,052 labeled sentences, respectively. For simplicity, we will refer to them as MI Augmented (Union) and MI Augmented (Intersection) datasets. 

%The MI Augmented (Union) and MI Augmented (Intersection) datasets contained 1,318,913 and 80,690 labeled sentences, respectively. 

%we computed the cosine similarity between the embeddings of the labeled sentences in the MI gold dataset and the unlabeled sentences among the responses in CounselChat and RED datasets and retrieved the unlabeled

%  and specifically the ones that can be rephrased into a more MI-adherent form

\section{MI Classifier}

We developed a classifier to automatically classify responses in distress-support dialogues into one of the 15 labels mentioned above. This is an important step that should be followed before rephrasing, since first it should identify the unfavourable responses types. For this purpose, we developed a classifier that consists of a representation network that uses the BERT architecture \cite{bert}, an attention layer that aggregates all hidden states at each time step, a hidden layer, and a softmax layer. We used the BERT-base architecture with 12 layers, 768 dimensions, 12 heads, and 110M parameters as the representation network. It was initialized with weights from RoBERTa \cite{roberta}. We trained three classifiers. The first one was trained on the smaller human-annotated MI dataset (MI Gold) taking 80\% of the data for training and leaving 10\% each for validation and testing. The other two were trained on the MI Augmented (Union) and MI Augmented (Intersection) datasets, leaving out the data used for validation and testing in the first case. In all cases, the optimal model was chosen based on average cross entropy loss calculated between the ground truth and predicted labels in the human-annotated validation set.

The classifiers trained on MI Gold, MI Augmented (Intersection), and MI Augmented (Union) datasets reported accuracies of 68.31\%, 67.13\%, and 73.44\% on the MI Gold test set, respectively. The reported accuracies on the MI Gold validation set were 67.08\%, 64.07\%, and 72.67\%, respectively for the three classifiers. Accordingly, the labels collected through the union of N-gram matching and cosine similarity-based methods improved the accuracy of the classifier by 8.33\% and 7.5\%, respectively on the validation and test sets compared to the accuracies reported when trained on the gold-labeled MI dataset. 

%the F1-scores (weighted average) on the MI gold test dataset are 68.07%, 65.85%, and 72.92%, respectively for the MI classifier trained on the MI Gold, MI Augmented (Intersection), and MI Augmented (Union) datasets

%Table \ref{table:classifier} shows the performance scores of the MI classifier when trained on gold-labeled and augmented MI datasets. According to the results, the labels collected through the union of the N-gram matching and cosine similarity-based methods, which accounts for $\approx$1.3M labels improved the accuracy of the classifier by 8.33\% and 7.5\%, respectively on the validation and test sets compared to the accuracies reported when trained on the gold-labeled MI dataset. 

%up to 72.67\% in the validation set and up to 73.44\% in the test set. This is an increase by 8.33\% and 7.5\%, respectively on the validation and test sets compared to the accuracies when trained on the gold-labeled MI dataset. 

%The confusion matrices associated with each of these classifiers are denoted in the appendices. 

\section{MI Rephraser}

% The second and final step after being able to detect any unfavourable response type is checking whether it is able to be rephrased into a form that is more MI adherent. 

After identifying the favourable and unfavourable response types, we can choose to omit the unfavourable responses or if possible, rephrase them into a more MI adherent form. A label pair that this rephrasing strategy can be applied directly are \textit{Advise without Permission} and \textit{Advise with Permission}. Through N-gram analysis, we could discover some N-gram patterns that are indicative of the label pair \textit{Advise without Permission} (e.g. \textit{You should}, \textit{You need to}, \textit{You musn't}) and \textit{Advise with Permission} (e.g. \textit{It maybe helpful to}, \textit{I wonder if you can}, \textit{You may want to consider}). These could be identified as style attributes that vary across the responses identified as \textit{Advise without Permission} and \textit{Advise with Permission}. Thus, given a response identified as \textit{Advise without Permission}, the goal of the rephraser would be to rephrase the response to be indicative of \textit{Advise with Permission}, without changing the semantic content of the response. 

%  preserving the semantic content of the original response

% Table \ref{tab:n-grams} shows some examples of these indicative N-grams. 

As mentioned in Section \ref{sec:lit_review}, this can be identified as a sub-task under the task of Text Style Transfer (TST). TST is formally defined as, given a target utterance $x'$ and the target discourse style attribute $a'$, model $p(x'|a, x)$, where $x$ is a given text carrying a source attribute value $a$. In our case, $x$ corresponds to the response identified as \textit{Advise without Permission}, $a$ corresponds to \textit{Advise without Permission}, and $a'$ corresponds to \textit{Advise with Permission}. 

% the source utterance 
% the source attribute 
% the target attribute 

\subsection{Pseudo-Parallel Corpora}

As discussed in Section \ref{sec:lit_review}, the most recent methods for TST involve data-driven deep learning models. The prerequisite for using such models is that there exist style-specific corpora for each style of interest, either parallel or nonparallel. With the human-annotated MI dataset, we are in possession of a non-parallel corpus containing 2,285 \textit{Advise without Permission} and 484 \textit{Advise with Permission} type of responses. With the MI Augmented (Union) dataset, we have 199,885 \textit{Advise without Permission} and 3,541 \textit{Advise with Permission} type of responses. Since creating parallel corpora consumes human labor and cost, using the above data, we decided to create pseudo-parallel corpora that contain pairs of \textit{Advise without Permission} and \textit{Advise with Permission} responses to train our rephrasers. We used two automatic methods to create these pseudo-parallel corpora: 1) Template-based replacement method; and 2) Retrieval method. 

% 193,674

% and use them to fine-tune two of the popular pre-trained language generation architectures Blender \cite{blender} and GPT3 \cite{gpt3}.

\subsubsection{Template-Based Replacement Method}

We used frequency-based N-gram analysis accompanied by human inspection to determine the linguistic templates that represent \textit{Advise with Permission} and \textit{Advise without Permission} responses. Table \ref{tab:template} shows some templates discovered for \textit{Advise without Permission} (on left) and \textit{Advise with Permission} (on right). In template-based replacement, if the algorithm detects any linguistic template on the left among the responses labeled as \textit{Advise without Permission}, it will randomly select a template from the right to replace it with, giving a pair of \textit{Advise without Permission} and \textit{Advise with Permission} responses that contain the same semantic content but differ in style. 

\begin{table}[ht!]
\small
\centering
%\begin{tabularx}{\linewidth}{X p{1cm} p{1cm} p{1cm}}
\begin{tabularx}{\linewidth}{p{3cm} | X}
%\begin{tabular}{l | l}
\toprule

\textbf{Advise without } &  \textbf{Advise with Permission}\\
\textbf{Permission} &  \\

\midrule

- \textit{You can} (verb) \underline{\hspace{0.5cm}} & - \textit{It maybe helpful to} (verb) \underline{\hspace{0.5cm}} \\
- \textit{You could} (verb) \underline{\hspace{0.5cm}} & - \textit{You may want to} (verb) \underline{\hspace{0.5cm}}\\
- \textit{You need to} (verb) \underline{\hspace{0.5cm}} & - \textit{I encourage you to} (verb) \underline{\hspace{0.5cm}}\\
- \textit{You should} (verb) \underline{\hspace{0.5cm}} & - \textit{Perhaps you can} (verb) \underline{\hspace{0.5cm}}\\
- (Verb) \underline{\hspace{0.5cm}} & - \underline{\hspace{0.5cm}}\textit{, if you would like.}\\

%- \textit{You can try to} (verb) \underline{\hspace{0.5cm}} & - \textit{It would be good idea to} (verb) \underline{\hspace{0.5cm}}\\
%- \textit{I think you should} (verb) \underline{\hspace{0.5cm}} & - \textit{It may be important to} (verb) \underline{\hspace{0.5cm}}\\
%- \textit{I suggest that you} (verb) \underline{\hspace{0.5cm}} & - \textit{I would encourage you to} (verb) \underline{\hspace{0.5cm}}\\
%- \textit{I suggest you} (verb) \underline{\hspace{0.5cm}} & - \textit{I wonder if you can} (verb) \underline{\hspace{0.5cm}}\\
%- \textit{Maybe you can} (verb) \underline{\hspace{0.5cm}} & - \textit{Maybe it is important to} (verb) \underline{\hspace{0.5cm}}\\
% \textit{Maybe you could} (verb) \underline{\hspace{0.5cm}} & - \textit{An option would be to} (verb) \underline{\hspace{0.5cm}}\\

%     & - \textit{You may want to consider} (present continuous form of the verb) \underline{\hspace{0.5cm}}\\
%      & - \textit{You may consider} (present continuous form of the verb) \underline{\hspace{0.5cm}}\\
%       & - \textit{I would recommend} (present continuous form of the verb) \underline{\hspace{0.5cm}}\\
%  & - \textit{I wonder if you can consider} (present continuous form the verb) \underline{\hspace{0.5cm}}\\

\bottomrule
%\end{tabular}
\end{tabularx}
\caption{Examples of templates corresponding to \textit{Advise without Permission} and \textit{Advise with Permission} responses. The full list is included in the appendices.}
\label{tab:template}
\vspace{-2mm}
\end{table}

% The list of phrases indicative of \textit{Advise without Permission} and \textit{Advise with Permission} are included in the appendices. 

% Table \ref{tab:template} shows the list of phrases indicative of \textit{Advise without Permission} (on the left) and \textit{Advise with Permission} (on the right). 

We constructed two pseudo-parallel corpora by applying this method to the MI Gold and MI Augmented (Union) datasets, which contained 2,285 and 199,885 responses labeled as \textit{Advise without Permission}, respectively. They respectively gave us 240 and 38,559 response pairs.

%The first one gave us 240 response pairs and the second gave us 38,559 response pairs.

% Then we handcraft some rules that can replace certain linguistic phrases present in \textit{Advise without Permission} responses with linguistic phrases indicative of \textit{Advise with Permission}.

\subsubsection{Retrieval Method} 

Given the non-parallel corpus containing \textit{Advise without Permission} and \textit{Advise with Permission} responses, we computed the semantic similarity between the \textit{Advise without Permission} and \textit{Advise with Permission} responses and retrieved the response pairs whose similarity is above a certain threshold. We used Sentence-BERT \cite{sbert} to generate embeddings of the two types of responses and compared them using cosine similarity. After manually inspecting a random subset of response pairs over a range of similarity thresholds, we chose 0.7 as the final threshold to determine the semantically similar response pairs. Similar to template-based replacement, we used this method to construct two pseudo-parallel corpora by applying the method to the gold-labeled and augmented-labeled MI datasets and obtained 104 and 54,956 response pairs, respectively. For simplicity, we will refer to the corpus constructed using the gold-labeled MI dataset as pseudo-parallel (PP) corpus and the corpus constructed using the augmented-labeled MI dataset as pseudo-parallel augmented (PPA) corpus. We used 80\% of the data from each of the corpora for training our rephrasers, and 10\% each for validation and testing. In section \ref{sec:human}, we gauge the quality of the above corpora using human ratings.   

% Figure \ref{fig:parallel-corpus} shows a summary of the different types of PP and PPA corpora developed. 

\subsection{Rephrasing Models}

% the PP and PPA corpora developed using template-based replacement and retrieval methods

Using the above corpora, we fine-tuned two pre-trained language generation architectures Blender \cite{blender} and GPT-3 \cite{gpt3}. Blender is a standard Seq2Seq transformer-based dialogue model. We used the 90M parameter version of Blender. Though it is a dialogue generation model, we used it mainly because it is pre-trained on Reddit discussions containing $\approx$1.5B comments and is already aware of the language constructs used in peer support. GPT-3 is a language model that utilizes standard transformer network having 175 billion parameters. We used the smallest but fastest version of GPT-3, Ada, to build our rephrasers. The main reason to use GPT-3 is that it has demonstrated strong few-shot learning capability on many text-based tasks. Both Blender and GPT-3 were fine-tuned on template-based, retrieval-based, and combined PP and PPA corpora.

%The latter is a standard Seq2Seq transformer-based empathetic open-domain chatbot. It is pre-trained on Reddit discussions containing $\approx$1.5B comments and fine-tuned on several smaller but focussed datasets. 

Prior work has shown large language models can perform various tasks given a clever prompt prepended to the input \cite{gpt3}. So, we developed two variations of Blender and GPT3 models by appending a generic prompt and an N-gram-based prompt to the end of the training data. In generic prompting, we simply appended the label \textbf{\textit{Advise with permission:}} to the end of the input text. In N-gram prompting, we detected if there is any N-gram that is indicative of \textit{Advise with permission} in the output text. If there is, we appended it to the end of the input text. Table \ref{tab:prompting} shows training examples with generic and N-gram-based prompts.

%  from template-based and retrieval-based PP and PPA corpora

\begin{table}[ht!]
\small
\centering
%\begin{tabularx}{\linewidth}{X p{1cm} p{1cm} p{1cm}}
\begin{tabularx}{\linewidth}{p{1cm} X}
\toprule

\multicolumn{2}{l}{\textbf{Training example with generic prompting:}} \vspace{0.5mm}\\
Input: & \textit{try to learn from your mistakes and meet some new people . \textbf{Advise with permission:}}\\
Output: & \textit{It may be important to try to learn from your mistakes and meet some new people.}\\

\midrule

\multicolumn{2}{l}{\textbf{Training example with N-gram based prompting:}} \vspace{0.5mm}\\
Input: & \textit{try to learn from your mistakes and meet some new people . \textbf{It may be important to:}}\\
Output: & \textit{\textbf{It may be important to} try to learn from your mistakes and meet some new people.}\\

\bottomrule
\end{tabularx}
\caption{Examples with generic and N-gram prompts.}
\label{tab:prompting}
\vspace{-5mm}
\end{table}

%\begin{figure}[ht!]
%     \centering
%     \subfloat[][Training example with generic prompting]{\includegraphics[width=0.9\linewidth]{Images/generic-prompt.png}\label{fig:generic}}
%     \qquad
%     \subfloat[][Training example with N-gram based prompting]{\includegraphics[width=0.9\linewidth]{Images/ngram-prompt.png}\label{fig:ngram}}
%     \caption{Training examples with generic and N-gram based prompts.}
%\label{fig:prompting}
%\end{figure}

Altogether we developed 10 different rephrasing models by fine-tuning Blender and GPT-3 on: 1) template-based PP and PPA corpora; 2) retrieval-based PP and PPA corpora; 3) combined template-based and retrieval-based PP and PPA corpora; 4) combined template and retrieval based PP and PPA corpora appending generic prompts; 5) combined template and retrieval based PP and PPA corpora appending N-gram prompts. Some examples of the rephrased output by these different models are shown in the appendices.

%Training details and some examples of rephrasings generated by the different models are included in the appendices.

%More details of these models including the hyperparameters used during training are included in the appendices. 

%\section{Evaluation}

%Evaluation performs an important role in measuring the effectiveness of the rephrasing strategy. We used a number of automatic and human evaluation metrics to evaluate the performance of the Blender and GPT3-based rephrasers. 

\section{Automatic Evaluation}

% \cite{tst_survey, tst_evaluation}

%  (in our case, \textit{Advise with Permission})

% related to text generation as well as text style transfer 

% , and maintain natural language fluency
% at the same time 

% We used a number of automatic metrics to evaluate the output from Blender and GPT3 based rephrasers across the above dimensions. 

A successful style-transferred output should be able to demonstrate the correct target style and at the same time preserve the semantic content of the original text \cite{tst_survey, tst_evaluation}. We refer to the first criterion as \textit{Style Transfer Strength} and the second as \textit{Semantic Similarity}. Automatic metrics used to evaluate text generation methods such as the BLEU score \cite{bleu}, ROUGE \cite{rouge}, METEOR \cite{meteor}, Word Mover Distance (WMD) \cite{wmd}, Character N-gram F-score (chrf) \cite{chrf}, BERTScore \cite{bertscore} and cosine similarity based on sentence embeddings \cite{sbert} are used in the literature to evaluate the semantic similarity between the original and the rephrased text. The Part-of-Speech distance \cite{pos}, a metric specific to TST, is also used to measure semantic similarity. Mir et al. \shortcite{mir} suggest deleting all attribute-related expressions in the text when applying these metrics to evaluate the output of TST tasks. Thus, before evaluation, we removed the style-specific phrases discovered during N-gram analysis from the input and output text.

% which denotes the noun difference between the original and transferred sentences, 

% and deep-learning based metrics such as the 

% In this way, these metrics can effectively evaluate the extent to which the style-independent content in the original text is preserved in the rephrased text. 

% , and then conduct the evaluations

%Thus, before evaluating the semantic similarity of the rephrased text generated by our rephrasers, we removed the style specific phrases discovered during N-gram analysis from the input and output text.

%We applied the above evaluation metrics to evaluate the semantic similarity between the original text and the rephrased text generated by the Blender and GPT-3 based rephrasers trained on PP and PPA datasets. They were tested on the PP test dataset. Before application, we removed the style specific phrases discovered during N-gram analysis from the input and output text.

% Metrics specific to Text Style Transfer (TST) such as 

% BERTScore = Analogously to common metrics, BERTScore computes a similarity score for each token in the candidate sentence with each token in the reference sentence. However, instead of exact matches, we compute token similarity using contextual embeddings.

%  The WMD distance measures the dissimilarity between two text documents as the minimum amount of distance that the embedded words of one document need to “travel” to reach the embedded words of another document. Retrieval based metric

% chrf: character n-gram F-score for automatic evaluation of machine translation output.

%\subsubsection{Style Transfer Strength} 

To evaluate the style transfer strength, most works use a style classifier to predict if the output conforms to the target style \cite{hu, li2018, prabhumoye}. We used the MI classifier trained on the MI Augmented (Union) dataset to compute the style transfer strength. It is calculated as the percentage of samples classified as \textit{Advise with Permission} out of all test samples.

\begin{table*}[ht!]
\small
\centering
%\begin{tabularx}{\linewidth}{X p{1cm} p{1cm} p{1cm}}
\begin{tabularx}{\linewidth}{X | r r | r r | r r | r r | r r }
\toprule

\textbf{Criteria} & \multicolumn{2}{l|}{\textbf{Template}} & \multicolumn{2}{l|}{\textbf{Retrieval}} & \multicolumn{2}{l|}{\textbf{Template +}} & \multicolumn{2}{l|}{\textbf{Template +}} & \multicolumn{2}{l}{\textbf{Template +}}\\

% &  &  & & & \multicolumn{2}{l|}{\textbf{+}} & \multicolumn{2}{l|}{\textbf{+}} & \multicolumn{2}{l}{\textbf{+}}\\

&  &  & & & \multicolumn{2}{l|}{\textbf{Retrieval}} & \multicolumn{2}{l|}{\textbf{Retrieval}} & \multicolumn{2}{l}{\textbf{Retrieval}}\\

&  &  & & & \multicolumn{2}{l|}{\textbf{}} & \multicolumn{2}{l|}{\textbf{(with generic}} & \multicolumn{2}{l}{\textbf{(with N-gram}}\\

&  &  & & & \multicolumn{2}{l|}{\textbf{}} & \multicolumn{2}{l|}{\textbf{prompting)}} & \multicolumn{2}{l}{\textbf{prompting)\vspace{1mm}}}\\

& \textbf{BB} & \textbf{GPT3} & \textbf{BB} & \textbf{GPT3} & \textbf{BB} & \textbf{GPT3} & \textbf{BB} & \textbf{GPT3} & \textbf{BB} & \textbf{GPT3}\\

\midrule

\multicolumn{11}{l}{\textbf{\vspace{0.5mm}Training dataset: PP\vspace{0.5mm}}}\\
%\multicolumn{11}{l}{\textbf{Trained on: PP; Tested on: PP\vspace{0.5mm}}}\\

BLEU-1 & 0.1315 & \textbf{0.3464} & 0.0787 & \textbf{0.1308} & 0.1429 & \textbf{0.2977} & \cellcolor{lightgreen} 0.1763 &\cellcolor{green} \textbf{0.3821} & 0.1585 &  \textbf{0.2751}\\

BLEU-2 & 0.0366 & \textbf{0.3225} & 0.0131 & \textbf{0.0501} & 0.0496 & \textbf{0.2671} & 0.0613 &  \cellcolor{green} \textbf{0.3556} & \cellcolor{lightgreen} 0.0677 & \textbf{0.2374}\\

BLEU-3 & \cellcolor{lightgreen} 0.0046 & \textbf{0.3120} & 0.0046 & \textbf{0.0328} & 0.0000 & \textbf{0.2543} & 0.0031 & \cellcolor{green} \textbf{0.3465} & 0.0000 & \textbf{0.2269}\\

BLEU-4 & \cellcolor{lightgreen} 0.0033 & \textbf{0.2994} & 0.0000 & \textbf{0.0326} & 0.0000 & \textbf{0.2262} & 0.0000 & \cellcolor{green} \textbf{0.3301} & 0.0000 & \textbf{0.2164}\\

ROUGE-L & 0.1760 & \textbf{0.5333} & 0.1176 & \textbf{0.1608} & 0.1843 & \textbf{0.4495} & \cellcolor{lightgreen} 0.2167 & \cellcolor{green} \textbf{0.5450} & 0.2135 & \textbf{0.4404}\\

METEOR & 0.1568 & \textbf{0.4622} & 0.0994 & \textbf{0.1323} & 0.1879 & \textbf{0.4210} & 0.2084 & \cellcolor{green} \textbf{0.5014} & \cellcolor{lightgreen} 0.2108 & \textbf{0.3726} \\

WMD $\downarrow$ & 1.0311 & \textbf{0.7068} & 1.1122 & \textbf{1.0800} & 1.0345 & \textbf{0.7928} & \cellcolor{lightgreen} 1.0073 & \cellcolor{green} \textbf{0.6746} & 1.0163 & \textbf{0.8447}\\

Chrf Score & 0.2690 & \textbf{0.5008} & 0.1678 & \textbf{0.2095} & 0.2690 & \textbf{0.4737} & \cellcolor{lightgreen} 0.3082 & \cellcolor{green} \textbf{0.5341} & 0.2955 & \textbf{0.4245}\\

BERTScore & 0.8656 & \cellcolor{green} \textbf{0.9138} & 0.8382 & \textbf{0.8658} & 0.8683 & \textbf{0.9048} & \cellcolor{lightgreen} 0.8821 & \textbf{0.9137} & 0.8693 & \textbf{0.9003}\\

POS dist. $\downarrow$ & 5.4771 & \cellcolor{green} \textbf{2.5523} & 9.8218 & \textbf{7.1482} & 5.8271 & \textbf{2.7042} & \cellcolor{lightgreen} 4.8378 & \textbf{2.5830} & 5.8854 & \textbf{3.6298}\\

Cos Similarity & 0.6116 & \cellcolor{green} \textbf{0.7524} & \textbf{0.4429} & 0.4291 & 0.6129 & \textbf{0.6516} & \cellcolor{lightgreen} 0.6918 & \textbf{0.7403} & \textbf{0.6571} & 0.6471 \\

Style Strength\vspace{0.5mm} & 29.41 & \textbf{73.53} & 0.00 & \textbf{47.06} & 38.24 & \cellcolor{lightgreen} \textbf{79.41} & \cellcolor{green} \textbf{94.12} & 61.76 & 23.53 & \textbf{58.82} \\

%\multicolumn{11}{l}{\textbf{Training dataset: PPA; Tested on: PP\vspace{0.5mm}}}\\
\multicolumn{11}{l}{\textbf{Training dataset: PPA\vspace{0.5mm}}}\\

BLEU-1 & 0.2039 & \cellcolor{green} \textbf{0.3751} & \textbf{0.2122} & 0.0987 & 0.2308 & \textbf{0.3229} & \cellcolor{lightgreen} 0.2588 & \textbf{0.3688} & 0.2021 & \textbf{0.3349}\\

BLEU-2 & 0.0913 & \cellcolor{green} \textbf{0.3456} & \textbf{0.1468} & 0.0263 & 0.1591 & \textbf{0.2836} & \cellcolor{lightgreen} 0.1849 & \textbf{0.3332} & 0.1455 & \textbf{0.3034}\\

BLEU-3 & 0.0031 & \cellcolor{green} \textbf{0.3352} & \textbf{0.1370} & 0.0172 & 0.1319 & \textbf{0.2725} & \cellcolor{lightgreen} 0.1536 & \textbf{0.3161} & 0.1239 & \textbf{0.2922}\\

BLEU-4 & 0.0000 & \cellcolor{green} \textbf{0.3217} & \textbf{0.1286} & 0.0069 & 0.1213 & \textbf{0.2536} & \cellcolor{lightgreen} 0.1437 & \textbf{0.2987} & 0.1169 & \textbf{0.2798}\\

ROUGE-L & 0.2642 & \cellcolor{green} \textbf{0.5363} & \textbf{0.2419} & 0.1216 & 0.2718 & \textbf{0.4467} & \cellcolor{lightgreen} 0.3016 & \textbf{0.5278} & 0.2352 & \textbf{0.5178}\\

METEOR & 0.3081 & \cellcolor{green} \textbf{0.4673} & \textbf{0.2436} & 0.1063 & 0.2932 & \textbf{0.4261} & \cellcolor{lightgreen} 0.3102 & \textbf{0.4607} & 0.2557 & \textbf{0.4381}\\

WMD $\downarrow$ & 0.9716 & \cellcolor{green} \textbf{0.6849} & \textbf{1.0069} & 1.1584 & \textbf{0.9451} & 0.9754 & \cellcolor{lightgreen} 0.9095 & \textbf{0.7258} & 1.0000 & \textbf{0.7927}\\

Chrf Score & 0.3758 & \textbf{0.5038} & \textbf{0.3550} & 0.1782 & 0.4005 & \textbf{0.4648} & \cellcolor{lightgreen} 0.4048 & \cellcolor{green} \textbf{0.5047} & 0.3672 & \textbf{0.4897}\\

BERTScore & 0.8770 & \textbf{0.9116} & \textbf{0.8748} & 0.8582 & 0.8795 & \textbf{0.9021} & \cellcolor{lightgreen} 0.8837 & \cellcolor{green} \textbf{0.9140} & 0.8700 & \textbf{0.9028}\\

POS dist. $\downarrow$ & 7.4745 & \cellcolor{green} \textbf{1.9593} & 8.0439 & \textbf{7.0396} & 6.9338 & \textbf{2.8695} & \cellcolor{lightgreen} 6.1747 & \textbf{2.6637} & 10.1620 & \textbf{3.0649}\\

Cos Similarity & \cellcolor{lightgreen} 0.6428 & \cellcolor{green} \textbf{0.7481} & \textbf{0.5910} & 0.4605 & 0.6277 & \textbf{0.6501} & 0.6303 & \textbf{0.7318} & 0.5717 & \textbf{0.6807} \\

Style Strength & \cellcolor{lightgreen} 73.53 & \cellcolor{green} \textbf{76.47} & \textbf{58.82} & 32.35 & \textbf{70.59} & 61.76 & \textbf{67.65} & 55.88 & \textbf{52.94} & 52.94 \\

\bottomrule

\end{tabularx}
\caption{Automatic evaluation results on PP test set. Under each method (Template, Retrieval etc.), the score of the rephraser that performs the best is made bold. The best score obtained for each of BB and GPT3-based rephrasers along each criteria is highlighted in green. Out of them, the best overall score is highlighted with a darker green.}
\label{table:automatic_eval_1}
\end{table*}

% Automatic evaluation results on the PP test set. Under each methodology used (Template, Retrieval etc.), the rephraser that performs the best out of BB and GPT3 is highlighted in bold. The best score obtained for each of BB and GPT3 based rephrasers along each human evaluation criteria are highlighted in green. Out of them, the best overall score is highlighted with a darker shade of green.

\begin{table*}[ht!]
\small
\centering
%\begin{tabularx}{\linewidth}{X p{1cm} p{1cm} p{1cm}}
\begin{tabularx}{\linewidth}{X | r r | r r | r r | r r | r r }
\toprule

\textbf{Criteria} & \multicolumn{2}{l|}{\textbf{Template}} & \multicolumn{2}{l|}{\textbf{Retrieval}} & \multicolumn{2}{l|}{\textbf{Template +}} & \multicolumn{2}{l|}{\textbf{Template +}} & \multicolumn{2}{l}{\textbf{Template +}}\\

% &  &  & & & \multicolumn{2}{l|}{\textbf{+}} & \multicolumn{2}{l|}{\textbf{+}} & \multicolumn{2}{l}{\textbf{+}}\\

 &  &  & & & \multicolumn{2}{l|}{\textbf{Retrieval}} & \multicolumn{2}{l|}{\textbf{Retrieval}} & \multicolumn{2}{l}{\textbf{Retrieval}}\\

 &  &  & & & \multicolumn{2}{l|}{\textbf{}} & \multicolumn{2}{l|}{\textbf{(with generic}} & \multicolumn{2}{l}{\textbf{(with N-gram}}\\

  &  &  & & & \multicolumn{2}{l|}{\textbf{}} & \multicolumn{2}{l|}{\textbf{prompting)}} & \multicolumn{2}{l}{\textbf{prompting)\vspace{1mm}}}\\

 & \textbf{BB} & \textbf{GPT3} & \textbf{BB} & \textbf{GPT3} & \textbf{BB} & \textbf{GPT3} & \textbf{BB} & \textbf{GPT3} & \textbf{BB} & \textbf{GPT3}\\

\midrule

\multicolumn{11}{l}{\textbf{Training dataset: PP; Tested on: PP\vspace{0.5mm}}}\\

Semantic Similarity (SS)\vspace{0.5mm} & 
1.74 & \cellcolor{green} \textbf{3.35} & 0.32 & \textbf{1.07} & 1.62 & \textbf{2.65} & \cellcolor{lightgreen} 2.49 & \textbf{2.72} & 1.88 & \textbf{2.31}\\

Style Transfer Strength (STS) & 2.78 & \cellcolor{lightgreen} \textbf{3.88} & 0.44 & \textbf{2.16} & 2.72 & \textbf{3.47} & \cellcolor{green}  \textbf{3.99} & 3.21 & 2.47 & \textbf{3.21}\\

\cline{2-11}
(Average of SS and STS) & 2.26 & \cellcolor{green} \textbf{3.62} & 0.54 & \textbf{1.62} & 2.17 & \textbf{3.06} & \cellcolor{lightgreen} \textbf{3.24} & 2.97 & 2.18 & \textbf{2.76}\vspace{1mm}\\

\multicolumn{11}{l}{\textbf{Training dataset: PP; Tested on: PPA\vspace{0.5mm}}}\\

Semantic Similarity (SS) & \textbf{2.07} & 0.69 & 0.79 & \textbf{0.94} & 2.22 & \textbf{2.60} & \cellcolor{lightgreen} 2.82 & \cellcolor{green} \textbf{2.87} & 2.10 & \textbf{2.50}\\

Style Transfer Strength (STS) & 2.51 & \cellcolor{lightgreen} \textbf{3.70} & 0.65 & \textbf{2.00} & 2.61 & \textbf{3.17} & \cellcolor{green} \textbf{3.96} & 3.14 & 2.26 & \textbf{3.02}\\

\cline{2-11}
(Average of SS and STS) & \textbf{2.29} & 2.20 & 0.72 & \textbf{1.47} & 2.42 & \textbf{2.89} & \cellcolor{green} \textbf{3.39} & \cellcolor{lightgreen} 3.01 & \textbf{3.23} & 2.76\\

\midrule

\multicolumn{11}{l}{\textbf{Training dataset: PPA; Tested on: PP\vspace{0.5mm}}}\\

Semantic Similarity (SS)\vspace{0.5mm} &  \cellcolor{lightgreen} 2.63 & \cellcolor{green} \textbf{3.19} & \textbf{1.21} & 0.81 & 1.69 & \textbf{2.57} & 1.74 & \textbf{2.53} & 1.21 & \textbf{2.32}\\

Style Transfer Strength (STS) &  \cellcolor{green} \textbf{3.94} & \cellcolor{lightgreen} 3.82 & \textbf{2.74} & 1.44 & 3.15 & \textbf{3.28} & 3.00 & \textbf{3.47} & 2.57 & \textbf{2.99}\\

\cline{2-11}
(Average of SS and STS) & \cellcolor{lightgreen} 3.29 & \cellcolor{green} \textbf{3.51} & \textbf{1.98} & 1.13 & 2.42 & \textbf{2.93} & 2.37 & \textbf{3.00} & 1.89 & \textbf{2.66}\vspace{1mm}\\

\multicolumn{11}{l}{\textbf{Training dataset: PPA; Tested on: PPA\vspace{0.5mm}}}\\

Semantic Similarity (SS) & \cellcolor{lightgreen} 2.78 & \cellcolor{green} \textbf{3.26} & \textbf{1.40} & 1.00 & 1.70 & \textbf{2.31} & 1.71 & \textbf{2.36} & 1.22 & \textbf{2.31} \\

Style Transfer Strength (STS) & \cellcolor{green} \textbf{3.92} & \cellcolor{lightgreen} 3.82 & \textbf{2.30} & 1.92 & 2.59 & \textbf{2.85} & 2.60 & \textbf{3.06} & 2.40 & \textbf{2.98} \\

\cline{2-11}
(Average of SS and STS) & \cellcolor{lightgreen} 3.35 & \cellcolor{green} \textbf{3.54} & \textbf{1.85} & 1.46 & \textbf{2.15} & \textbf{2.58} & 2.16 & \textbf{2.71} & 1.81 & \textbf{2.65} \\

\bottomrule
\end{tabularx}
\caption{Results of human evaluation. Under each methodology (Template, Retrieval etc.), the score of the rephraser that performs the best is highlighted in bold. The best score obtained for each of BB and GPT3-based rephrasers along each criteria is highlighted in green. Out of them, the best overall score is highlighted with a darker green.}
\label{table:human_eval}
\end{table*}

% The scores are computed based on the average rating of two workers. The ratings spanned from 0 to 4 with 4 being the best.

%  out of BB and GPT3 

% BB refers to the BlenderBot-based rephraser. 

%\begin{itemize}

%\item \textbf{Which type of backbone model work generally better for the task of rephrasing \textit{Advise without Permission} responses to \textit{Advise with Permission}?} 

%\end{itemize}

%\begin{itemize}

%\item \textbf{Does data augmentation improve the performance scores of Blender and GPT-3 based rephrasers?} 

%\end{itemize}

%To do

%\begin{itemize}

%\item \textbf{Does combining pseudo-parallel corpora generated using template and retrieval-based methods improve the performance scores of Blender and GPT-3 based rephrasers compared to using these corpora alone?}

%\end{itemize}

%To do

%\begin{itemize}

%\item \textbf{Does generic prompting help to improve the performance scores across Blender and GPT-3 based rephrasers?}

%\end{itemize}

%In Blender-based rephrasers, all the automatic evaluation metrics were improved after incorporating generic prompting, except for the style transfer strength when trained on the PPA corpus. 

Table \ref{table:automatic_eval_1} shows the results of automatic evaluation of the rephrasers on the combined PP test dataset, which contains data from both template and retrieval-based PP test sets. Accordingly, GPT3-based rephrasers show better performance compared to Blender-based rephrasers in 85\% of the time across the metrics. It could also be observed that data augmentation improves the scores across most metrics irrespective of the backbone model used. Combining the pseudo-parallel corpora obtained from template-based and retrieval-based methods could improve the performance scores of Blender-based rephrasers across most automatic metrics. But GPT-3 based rephrasers trained only on template-based pseudo-parallel data seem to achieve better scores across almost all the metrics when compared to those trained on retrieval-based and combined corpora.

%Table \ref{table:automatic_eval_1} shows the results of automatic evaluation on the PP test dataset. According to the results, GPT3-based rephrasers show better performance compared to Blender-based rephrasers in 85\% of the time across the metrics. It could also be observed that data augmentation generally help to improve the scores across most metrics irrespective of the backbone model used. Combining the pseudo-parallel corpora obtained from template-based and retrieval-based methods could improve the performance scores of Blender-based rephrasers in 8 out of 12 and 7 out of 12 automatic metrics when trained on PP and PPA datasets, respectively. However, GPT-3 based rephrasers trained only on template based pseudo-parallel data seem to achieve better scores across 11 out of 12 and 12 out of 12 metrics when trained on PP and PPA datasets, repectively, when compared to those trained on retrieval-based and combined corpora.

% As denoted by light green in Table \ref{table:automatic_eval_1}, 
% As denoted by dark green, 

Blender-based rephrasers that incorporated generic prompting ranked the best across most metrics over all the other Blender-based rephrasers. With the smaller PP training corpus, the GPT-3-based rephraser that incorporated generic prompting ranked the best across most metrics. But with the larger PPA training corpus, the GPT-3 based rephraser that was trained on simple template-replaced pseudo-parallel corpora ranked the best across most automatic metrics.

\section{Human Evaluation}
\label{sec:human}

%Adapted from the literature on "Text Style Transfer",  according to the above criteria

Similar to automatic evaluation, we used two human evaluation criteria to rate the rephrased sentences. The first is how close the rephrased sentence is to \textit{Advise with permission} (Style transfer strength). The second is to what extent the rephrased sentence preserves the context/meaning of the original sentence (Semantic similarity).

We used the UpWork crowdsourcing platform ({\url{www.upwork.com}}) and recruited four professional counselors to rate the rephrased sentences. Given the original \textit{Advise without Permission} sentence and a list of rephrased sentences generated by the 10 different rephrasers, we asked two questions from the counselors: 1) \textit{Is the rephrased sentence indicative of Advise with permission?}; and 2) \textit{Does the rephrased sentence preserve the original context?} The counselors were asked to answer these questions by indicating a rating on a Likert scale ranging from 0 (\textit{Not at all}) to 4 (\textit{Yes it is}). Along with the rephrased sentences, we also presented them the corresponding \textit{Advise with permission} sentence obtained from the pseudo-parallel corpora in order to gauge the quality of the corpora used for training. The sentences to be rated were presented to them in a random order to reduce bias. 

As the combined PP test corpus developed on the MI Gold dataset is small (only 34 samples), we used 200 randomly selected samples from the combined PPA test corpus developed on the augmented MI dataset to be rated by the human workers. This was to verify the trend of results reported on the PP test corpus. We bundled 9 randomly selected test cases in one batch and allocated two workers to rate each batch. Results were calculated based on the average rating given by the two workers. Following Adiwardana et al. \shortcite{meena} we also calculated the average of style transfer strength and semantic similarity ratings to obtain a single score. We computed the inter-rater agreement based on weighted Kappa that uses Fleiss-Cohen weights \cite{kappa} and the scores were 0.5870 (moderate agreement) and 0.6933 (substantial agreement) for style transfer strength and semantic similarity, respectively.

%, indicating moderate agreement on style transfer strength and substantial agreement on semantic similarity. 

% between the workers 

Table \ref{table:human_eval} shows the results of the human evaluation experiment. According to the results, GPT3-based rephrasers win over Blender-based rephrasers 70\% and 85\% of the time along style transfer and semantic similarity dimensions, respectively. And when it comes to the smaller PP training corpus, using generic prompting during training increases the scores across most cases. But when it comes to the larger PPA corpus, simply training the rephrasers with template-replaced pseudo-parallel pairs gives the best results irrespective of the underlying backbone model. 

%  constructed from the gold-labeled MI dataset
%  constructed from the augmented MI dataset

%These observations were quite consistent with the automatic evaluation results.  

The average ratings obtained for \textit{style transfer strength} and \textit{semantic similarity} for sentence pairs in the PP test corpus were 3.21 and 3.16, respectively. The sentence pairs in the PPA test corpus scored 3.12 and 2.69 in the above two dimensions, respectively. The average ratings being close to 3 with most of them being above 3 suggests that the training corpora used are of substantial quality. 

\section{Discussion}

In this paper, we presented an example on how distress-consoling responses could be boosted with MI strategy. For this, we first developed a classifier that can identify favourable and unfavourable response types as defined by the MITI code. Then we narrowed our focus to the MI non-adherent response type \textit{Advise without Permission} and developed several rephrasers that can rephrase \textit{Advise without Permission} responses into MI adherent response type \textit{Advise with Permission}. As curating human written rephrasings was costly, we used templated-based replacement and retrieval methods to create pseudo-parallel corpora from gold-labeled and augmented-labeled MI datasets that contained responses from Reddit and CounselChat platforms. We used this data to train several Blender and GPT3-based rephrasers. We also used generic and N-gram-based prompts to see if prompting can improve the rephrasers' performance.  

% developing a parallel corpus that contained \textit{Advise without Permission} responses directly rephrased by humans into \textit{Advise with Permission} responses was costly, 

Automatic as well as human evaluation results suggested fine-tuning on GPT3 gives better results in rephrasing \textit{Advise without permission} responses into \textit{Advise with permission}. Data augmentation techniques we used by expanding the MITI labels using N-gram-based matching and similarity-based retrieval improved the performance of the MI classifier as well as the Blender and GPT3-based rephrasers. The results also suggested when the training datasets are small, the use of generic prompting can enable the rephrasing models to produce better results across style transfer and semantic similarity dimensions. But if you are dealing with large datasets (in our case through data augmentation), pseudo-parallel data generated through simpler methods such as template-based replacement can enable the models to generate substantially good rephrasings closer to the required style and semantically similar to the original sentence.

% The pseudo-parallel corpora created on augmented labeled dataset improved the scores across both Blender and GPT3-based rephrasers. 

% data augmentation methods are used to make the training datasets larger

%To do

In the future, we hope to develop a chatbot that can respond to psychological distress using the RED dataset that contain dialogues curated from several mental health-related subreddits. Then we hope to improve the responses generated by this chatbot by applying MI boosting at two different levels: one at the data level; and the other at the model level. At data level boosting, we hope to apply the MI classifier and automatically label the responses in the training data itself. By doing so, we will be able to rephrase the MI non-adherent responses such as \textit{Advise without Permission} into more MI-adherent responses and omit the other unfavourable responses from the training data. The MI-boosted training data can then be used to train the chatbot. At model-level boosting, a similar methodology can be applied at the level the chatbot is decoding responses (e.g. beam search). Not only generative chatbots but also retrieval-based chatbots could be benefited from this methodology. 

%The application of this work is not limited to improving chatbot responses for distress consolation. This could also be applied for suggesting better responses when peers untrained in the practice of counseling attempt to respond to distress-related posts on peer support platforms such as Reddit.

%For example, when the responses are ranked at the model-level, higher ranking scores could be allocated for .  

%When evaluating such rephrasings using human evaluators, the questions posed to them could be altered for example,  

%\newpage

\section{Limitations}

% can contribute to incorrectly predicting responses as \textit{Advise without Permission}. The data augmentation mechanisms we used in this work may add noise for the pseudo-parallel corpora and thus may have an effect on the performance of the rephrasers. 

% The inaccuracies of the MI classifier as well as the various data augmentation methods used may add noise and can result in reduced performance. 

%The current work serves as a proof of concept of how distress support responses could be boosted with MI strategy. We only apply this to the case of rephrasing responses detected as \textit{Advise without Permission} into \textit{Advise with Permission}. 

Certain parts of our proposed methodology, for example, template-based replacement and n-gram-based prompting are applicable only when style-specific linguistic attributes could be identified between the source and the target text. And due to the cost of human labor and the lack of publicly available client-therapist dialogues, the sample size drawn in the study is small and thus may have an impact on the conclusions drawn. Our methods have only been tested for the English language. But we believe similar methods could be applied to other languages given they have unparallel corpora tagged with \textit{Advise without Permission} and \textit{Advise with Permission} labels. The rephrasing methods described in this paper are tested for short sentences with a maximum sentence length of 98 tokens. Thus, the scalability of these methods for long text still remains to be tested. 

When testing the rephrasers, there are some combinations that could be tried other than the ones already tested. For example, more models can be fine-tuned and tested separately on template-replaced and retrieval-based PP and PPA corpora but incorporating generic and N-gram prompting. In this work, we first combined these two types of corpora before attempting prompting since we could observe better performance on Blender when the corpora were combined. 

In order to have more data, we combined the \textit{Advise with Permission} and \textit{Advise without Permission} responses present in CounselChat and RED datasets. But studies show that there are differences in the language used by counselors and peers \cite{lahnala, mousavi}. So, there can be linguistic differences between the same type of response in CounselChat and RED datasets. Future work should attempt to identify these differences and ideally rephrase the responses given by peers to reflect the language of the counselors. 

%PP and PPA corpora incorporating generic and N-gram prompting and using PP and PPA corpora generated only through retrieval methods incorporating generic and N-gram prompting. Also, 

% We first combined the PP and PPA corpora generated with template-replaced and retrieval methods since we could see better performance when Blender was fine-tuned on the combined corpora compared to Blender models fine-tuned on template-replaced and retrieval-based PP and PPA corpora separately. 

\section{Ethics Statement}

\textbf{Data Curation:} Only publicly available data in Reddit and CounselChat websites were used in this work. Analysis of posts on websites such as Reddit is considered "fair play" since individuals are anonymous and users are aware their responses remain archived on the site unless explicitly deleted. It is also stated in Reddit's privacy policy that it allows third parties to access public Reddit content. \footnote{\url{www.redditinc.com/policies/privacy-policy-october-15-2020}} Also, Reddit's data is already widely available in larger dumps such as Pushshift \cite{baumgartner2020pushshift}. Even though the policies allow it, it should be thoroughly noted that this data contains sensitive information. Thus, we adhere to the guidelines suggested by Benton et al. \shortcite{benton} for working with social media data in health research, and share only anonymized and paraphrased excerpts from the dataset so that it is not possible to recover usernames through a web search with the verbatim post text. In addition, references to usernames as well as URLs are removed from dialogue content for de-identification. 

%Analysis of posts of a website such as Reddit is likely considered ``fair play" as individuals are anonymous and users are aware that their responses remain archived on the site unless they explicitly delete them. The Reddit privacy policy also states it allows third parties to access public Reddit content through the Reddit API and other similar technologies. \footnote{\url{www.redditinc.com/policies/privacy-policy-october-15-2020}} Reddit's data is already widely available in larger dumps such as Pushshift \cite{baumgartner2020pushshift}. We collected only publicly available data in Reddit and the curation process did not involve any intervention or interaction with the Reddit users. The CounselChat dataset is also available publicly. But Fiesler and Proferes \shortcite{fiesler} in a study on user perceptions on social media research ethics empahsizes some potential harms that can be caused due to social computing research because internet users rarely read or could fully understand website terms and conditions. Since this dataset in particular contains sensitive information, we adhere to the guidelines suggested by Benton et al. \shortcite{benton} for working with social media data in health research, and share only anonymized and paraphrased excerpts from the dataset so that it is not possible to recover usernames through a web search with the verbatim post text. In addition, references to usernames as well as URLs are removed from dialogue content for de-identification. 

\textbf{Human Evaluation:} The human raters recruited from the crowdsourcing platform, UpWork, were all trained in the practice of counseling. Since the methods were tested on English-only text, we recruited workers who had professional competency in the English language. We paid them \$10 for evaluating each batch of rephrased sentences that required on average $\approx$30 minutes to complete. Thus, the amount paid to the human raters was $\approx$2.75 times above the US minimum wage of \$7.25 per hour. We also paid an extra \$2 as a bonus per each batch for workers who obtained an above-average agreement with the other worker who rated the same batch. 

\textbf{Chatbots for Distress-Consolation:} One of the main applications of the proposed methodology is boosting chatbot responses for distress consolation with motivational interviewing strategy. Using chatbots for distress consolation or other mental health interventions has raised ethical concerns among many \cite{lanteigne,montemayor,tatman}. However, chatbots that intervene in mental health-related matters have already been developed and have been quite popular for a while. Some examples are SimSensei \cite{simsei}, Dipsy \cite{dipsy}, Woebot (\url{woebothealth.com}), and Wysa (\url{www.wysa.io}). Czerwinski et al. \shortcite{microsoft} state, \textit{About 1 billion people globally are affected by mental disorders; a scalable solution such as an AI therapist could be a huge boon}. The current technology to develop such chatbots rely heavily on deep learning and pre-trained language models. But due to the inherently unpredictable nature of these models, they pose a threat of delivering unfavourable responses when such chatbots are used for distress consolation. We believe the methodology we suggest in this work can help them become more reliable and fail-safe by adhering to the motivational interviewing strategy, a guiding style of communication heavily practiced in psychotherapy. However, since the unfavourable response detection and rephrasing methods still rely on neural network models, the artifacts produced in this paper should be used for research purposes only and real-world deployment of them should be done under human supervision.

% Emma \cite{emma}, 

%and the artifacts produced in this paper 

%And the work we propose here is to make them more reliable and fail-safe. 

\bibliography{anthology,custom}

\begin{thebibliography}{59}
\expandafter\ifx\csname natexlab\endcsname\relax\def\natexlab#1{#1}\fi

\bibitem[{Adiwardana et~al.(2020)Adiwardana, Luong, So, Hall, Fiedel,
  Thoppilan, Yang, Kulshreshtha, Nemade, Lu et~al.}]{meena}
Daniel Adiwardana, Minh-Thang Luong, David~R So, Jamie Hall, Noah Fiedel, Romal
  Thoppilan, Zi~Yang, Apoorv Kulshreshtha, Gaurav Nemade, Yifeng Lu, et~al.
  2020.
\newblock Towards a human-like open-domain chatbot.
\newblock \emph{arXiv preprint arXiv:2001.09977}.

\bibitem[{Alambo et~al.(2019)Alambo, Gaur, Lokala, Kursuncu, Thirunarayan,
  Gyrard, Sheth, Welton, and Pathak}]{suicide}
Amanuel Alambo, Manas Gaur, Usha Lokala, Ugur Kursuncu, Krishnaprasad
  Thirunarayan, Amelie Gyrard, Amit Sheth, Randon~S Welton, and Jyotishman
  Pathak. 2019.
\newblock Question answering for suicide risk assessment using reddit.
\newblock In \emph{2019 IEEE 13th International Conference on Semantic
  Computing (ICSC)}, pages 468--473. IEEE.

\bibitem[{Banerjee and Lavie(2005)}]{meteor}
Satanjeev Banerjee and Alon Lavie. 2005.
\newblock \href {https://aclanthology.org/W05-0909} {{METEOR}: An automatic
  metric for {MT} evaluation with improved correlation with human judgments}.
\newblock In \emph{Proceedings of the {ACL} Workshop on Intrinsic and Extrinsic
  Evaluation Measures for Machine Translation and/or Summarization}, pages
  65--72, Ann Arbor, Michigan. Association for Computational Linguistics.

\bibitem[{Baumgartner et~al.(2020)Baumgartner, Zannettou, Keegan, Squire, and
  Blackburn}]{baumgartner2020pushshift}
Jason Baumgartner, Savvas Zannettou, Brian Keegan, Megan Squire, and Jeremy
  Blackburn. 2020.
\newblock The pushshift reddit dataset.
\newblock \emph{Proceedings of the International AAAI Conference on Web and
  Social Media}, 14(1):830--839.

\bibitem[{Benton et~al.(2017)Benton, Coppersmith, and Dredze}]{benton}
Adrian Benton, Glen Coppersmith, and Mark Dredze. 2017.
\newblock Ethical research protocols for social media health research.
\newblock In \emph{Proceedings of the First ACL Workshop on Ethics in Natural
  Language Processing}, pages 94--102.

\bibitem[{Bowman et~al.(2015)Bowman, Angeli, Potts, and
  Manning}]{A_large_annotated_corpus_for_learning_natural_language_inference}
Samuel~R. Bowman, Gabor Angeli, Christopher Potts, and Christopher~D. Manning.
  2015.
\newblock A large annotated corpus for learning natural language inference.
\newblock In \emph{Proceedings of the 2015 Conference on Empirical Methods in
  Natural Language Processing}, pages 632--642, Lisbon, Portugal. Association
  for Computational Linguistics.

\bibitem[{Brown et~al.(2020)Brown, Mann, Ryder, Subbiah, Kaplan, Dhariwal,
  Neelakantan, Shyam, Sastry, Askell, Agarwal, Herbert-Voss, Krueger, Henighan,
  Child, Ramesh, Ziegler, Wu, Winter, Hesse, Chen, Sigler, Litwin, Gray, Chess,
  Clark, Berner, McCandlish, Radford, Sutskever, and Amodei}]{gpt3}
Tom Brown, Benjamin Mann, Nick Ryder, Melanie Subbiah, Jared~D Kaplan, Prafulla
  Dhariwal, Arvind Neelakantan, Pranav Shyam, Girish Sastry, Amanda Askell,
  Sandhini Agarwal, Ariel Herbert-Voss, Gretchen Krueger, Tom Henighan, Rewon
  Child, Aditya Ramesh, Daniel Ziegler, Jeffrey Wu, Clemens Winter, Chris
  Hesse, Mark Chen, Eric Sigler, Mateusz Litwin, Scott Gray, Benjamin Chess,
  Jack Clark, Christopher Berner, Sam McCandlish, Alec Radford, Ilya Sutskever,
  and Dario Amodei. 2020.
\newblock \href
  {https://proceedings.neurips.cc/paper/2020/file/1457c0d6bfcb4967418bfb8ac142f64a-Paper.pdf}
  {Language models are few-shot learners}.
\newblock In \emph{Advances in Neural Information Processing Systems},
  volume~33, pages 1877--1901. Curran Associates, Inc.

\bibitem[{Cer et~al.(2017)Cer, Diab, Agirre, Lopez-Gazpio, and
  Specia}]{SemEval-2017_Task1_Semantic_Textual_Similarity_Multilingual_and_Crosslingual_Focused_Evaluation}
Daniel Cer, Mona Diab, Eneko Agirre, I{\~n}igo Lopez-Gazpio, and Lucia Specia.
  2017.
\newblock {S}em{E}val-2017 task 1: Semantic textual similarity multilingual and
  crosslingual focused evaluation.
\newblock In \emph{Proceedings of the 11th International Workshop on Semantic
  Evaluation ({S}em{E}val-2017)}, pages 1--14, Vancouver, Canada. Association
  for Computational Linguistics.

\bibitem[{Czerwinski et~al.(2021)Czerwinski, Hernandez, and McDuff}]{microsoft}
Mary Czerwinski, Javier Hernandez, and Daniel McDuff. 2021.
\newblock \href {https://doi.org/10.1109/MSPEC.2021.9423818} {Building an ai
  that feels: Ai systems with emotional intelligence could learn faster and be
  more helpful}.
\newblock \emph{IEEE Spectrum}, 58(5):32--38.

\bibitem[{De~Choudhury and De(2014)}]{de2014mental}
Munmun De~Choudhury and Sushovan De. 2014.
\newblock Mental health discourse on reddit: Self-disclosure, social support,
  and anonymity.
\newblock In \emph{Eighth international AAAI conference on weblogs and social
  media}.

\bibitem[{DeVault et~al.(2014)DeVault, Artstein, Benn, Dey, Fast, Gainer,
  Georgila, Gratch, Hartholt, Lhommet et~al.}]{simsei}
David DeVault, Ron Artstein, Grace Benn, Teresa Dey, Ed~Fast, Alesia Gainer,
  Kallirroi Georgila, Jon Gratch, Arno Hartholt, Margaux Lhommet, et~al. 2014.
\newblock Simsensei kiosk: A virtual human interviewer for healthcare decision
  support.
\newblock In \emph{Proceedings of the 2014 international conference on
  Autonomous agents and multi-agent systems}, pages 1061--1068.

\bibitem[{Devlin et~al.(2019)Devlin, Chang, Lee, and Toutanova}]{bert}
Jacob Devlin, Ming-Wei Chang, Kenton Lee, and Kristina Toutanova. 2019.
\newblock {BERT}: Pre-training of deep bidirectional transformers for language
  understanding.
\newblock In \emph{Proceedings of the 2019 Conference of the North {A}merican
  Chapter of the Association for Computational Linguistics: Human Language
  Technologies, Volume 1 (Long and Short Papers)}, pages 4171--4186,
  Minneapolis, Minnesota. Association for Computational Linguistics.

\bibitem[{Fitzpatrick et~al.(2017)Fitzpatrick, Darcy, and Vierhile}]{woebot}
Kathleen~Kara Fitzpatrick, Alison Darcy, and Molly Vierhile. 2017.
\newblock Delivering cognitive behavior therapy to young adults with symptoms
  of depression and anxiety using a fully automated conversational agent
  (woebot): a randomized controlled trial.
\newblock \emph{JMIR mental health}, 4(2):e7785.

\bibitem[{Fu et~al.(2018)Fu, Tan, Peng, Zhao, and Yan}]{tst_evaluation}
Zhenxin Fu, Xiaoye Tan, Nanyun Peng, Dongyan Zhao, and Rui Yan. 2018.
\newblock Style transfer in text: Exploration and evaluation.
\newblock In \emph{Proceedings of the AAAI Conference on Artificial
  Intelligence}, volume~32.

\bibitem[{Gan et~al.(2017)Gan, Gan, He, Gao, and Deng}]{gan2017}
Chuang Gan, Zhe Gan, Xiaodong He, Jianfeng Gao, and Li~Deng. 2017.
\newblock Stylenet: Generating attractive visual captions with styles.
\newblock In \emph{Proceedings of the IEEE Conference on Computer Vision and
  Pattern Recognition}, pages 3137--3146.

\bibitem[{Gaume et~al.(2009)Gaume, Gmel, Faouzi, and Daeppen}]{gaume}
Jacques Gaume, Gerhard Gmel, Mohamed Faouzi, and Jean-Bernard Daeppen. 2009.
\newblock Counselor skill influences outcomes of brief motivational
  interventions.
\newblock \emph{Journal of substance abuse treatment}, 37(2):151--159.

\bibitem[{Hendrycks and Gimpel(2016)}]{hendrycks2016gaussian}
Dan Hendrycks and Kevin Gimpel. 2016.
\newblock Gaussian error linear units (gelus).
\newblock \emph{arXiv preprint arXiv:1606.08415}.

\bibitem[{Hu et~al.(2017)Hu, Yang, Liang, Salakhutdinov, and Xing}]{hu}
Zhiting Hu, Zichao Yang, Xiaodan Liang, Ruslan Salakhutdinov, and Eric~P Xing.
  2017.
\newblock Toward controlled generation of text.
\newblock In \emph{International conference on machine learning}, pages
  1587--1596. PMLR.

\bibitem[{Huang et~al.(2020)Huang, Zhu, Xiong, Zhang, Hu, and Xu}]{yahoo}
Yufang Huang, Wentao Zhu, Deyi Xiong, Yiye Zhang, Changjian Hu, and Feiyu Xu.
  2020.
\newblock Cycle-consistent adversarial autoencoders for unsupervised text style
  transfer.
\newblock \emph{arXiv preprint arXiv:2010.00735}.

\bibitem[{Inkster et~al.(2018)Inkster, Sarda, Subramanian et~al.}]{wysa}
Becky Inkster, Shubhankar Sarda, Vinod Subramanian, et~al. 2018.
\newblock An empathy-driven, conversational artificial intelligence agent
  (wysa) for digital mental well-being: real-world data evaluation
  mixed-methods study.
\newblock \emph{JMIR mHealth and uHealth}, 6(11):e12106.

\bibitem[{Jin et~al.(2022)Jin, Jin, Hu, Vechtomova, and Mihalcea}]{tst_survey}
Di~Jin, Zhijing Jin, Zhiting Hu, Olga Vechtomova, and Rada Mihalcea. 2022.
\newblock Deep learning for text style transfer: A survey.
\newblock \emph{Computational Linguistics}, 48(1):155--205.

\bibitem[{Jin et~al.(2019)Jin, Jin, Mueller, Matthews, and Santus}]{jin2019}
Zhijing Jin, Di~Jin, Jonas Mueller, Nicholas Matthews, and Enrico Santus. 2019.
\newblock {IM}a{T}: Unsupervised text attribute transfer via iterative matching
  and translation.
\newblock In \emph{Proceedings of the 2019 Conference on Empirical Methods in
  Natural Language Processing and the 9th International Joint Conference on
  Natural Language Processing (EMNLP-IJCNLP)}, pages 3097--3109, Hong Kong,
  China. Association for Computational Linguistics.

\bibitem[{Kusner et~al.(2015)Kusner, Sun, Kolkin, and Weinberger}]{wmd}
Matt Kusner, Yu~Sun, Nicholas Kolkin, and Kilian Weinberger. 2015.
\newblock From word embeddings to document distances.
\newblock In \emph{International conference on machine learning}, pages
  957--966. PMLR.

\bibitem[{Lahnala et~al.(2021)Lahnala, Zhao, Welch, Kummerfeld, An, Resnicow,
  Mihalcea, and P{\'e}rez-Rosas}]{lahnala}
Allison Lahnala, Yuntian Zhao, Charles Welch, Jonathan~K. Kummerfeld,
  Lawrence~C An, Kenneth Resnicow, Rada Mihalcea, and Ver{\'o}nica
  P{\'e}rez-Rosas. 2021.
\newblock \href {https://doi.org/10.18653/v1/2021.findings-acl.392} {Exploring
  self-identified counseling expertise in online support forums}.
\newblock In \emph{Findings of the Association for Computational Linguistics:
  ACL-IJCNLP 2021}, pages 4467--4480, Online. Association for Computational
  Lingfgfggftzr666757tl.uistics.

\bibitem[{Lanteigne(2019)}]{lanteigne}
Camylle Lanteigne. 2019.
\newblock Social robots and empathy: The harmful effects of always getting what
  we want.

\bibitem[{Li et~al.(2018)Li, Jia, He, and Liang}]{li2018}
Juncen Li, Robin Jia, He~He, and Percy Liang. 2018.
\newblock Delete, retrieve, generate: a simple approach to sentiment and style
  transfer.
\newblock In \emph{Proceedings of the 2018 Conference of the North {A}merican
  Chapter of the Association for Computational Linguistics: Human Language
  Technologies, Volume 1 (Long Papers)}, pages 1865--1874. Association for
  Computational Linguistics.

\bibitem[{Lin and Och(2004)}]{rouge}
Chin-Yew Lin and Franz~Josef Och. 2004.
\newblock \href {https://doi.org/10.3115/1218955.1219032} {Automatic evaluation
  of machine translation quality using longest common subsequence and
  skip-bigram statistics}.
\newblock In \emph{Proceedings of the 42nd Annual Meeting of the Association
  for Computational Linguistics ({ACL}-04)}, pages 605--612, Barcelona, Spain.

\bibitem[{Liu et~al.(2022)Liu, Gao, Jia, Xu, and Vosoughi}]{liu2022}
Ruibo Liu, Chongyang Gao, Chenyan Jia, Guangxuan Xu, and Soroush Vosoughi.
  2022.
\newblock Non-parallel text style transfer with self-parallel supervision.
\newblock \emph{arXiv preprint arXiv:2204.08123}.

\bibitem[{Liu et~al.(2019)Liu, Ott, Goyal, Du, Joshi, Chen, Levy, Lewis,
  Zettlemoyer, and Stoyanov}]{roberta}
Yinhan Liu, Myle Ott, Naman Goyal, Jingfei Du, Mandar Joshi, Danqi Chen, Omer
  Levy, Mike Lewis, Luke Zettlemoyer, and Veselin Stoyanov. 2019.
\newblock Roberta: A robustly optimized bert pretraining approach.
\newblock \emph{arXiv preprint arXiv:1907.11692}.

\bibitem[{Madaan et~al.(2020)Madaan, Setlur, Parekh, Poczos, Neubig, Yang,
  Salakhutdinov, Black, and Prabhumoye}]{madaan2020}
Aman Madaan, Amrith Setlur, Tanmay Parekh, Barnabas Poczos, Graham Neubig,
  Yiming Yang, Ruslan Salakhutdinov, Alan~W Black, and Shrimai Prabhumoye.
  2020.
\newblock Politeness transfer: A tag and generate approach.
\newblock In \emph{Proceedings of the 58th Annual Meeting of the Association
  for Computational Linguistics}, pages 1869--1881, Online. Association for
  Computational Linguistics.

\bibitem[{Mairesse and Walker(2011)}]{mairesse2011}
Fran{\c{c}}ois Mairesse and Marilyn~A Walker. 2011.
\newblock Controlling user perceptions of linguistic style: Trainable
  generation of personality traits.
\newblock \emph{Computational Linguistics}, 37(3):455--488.

\bibitem[{Miller et~al.(2017)Miller, Feng, Batra, Bordes, Fisch, Lu, Parikh,
  and Weston}]{parlai}
Alexander Miller, Will Feng, Dhruv Batra, Antoine Bordes, Adam Fisch, Jiasen
  Lu, Devi Parikh, and Jason Weston. 2017.
\newblock \href {https://doi.org/10.18653/v1/D17-2014} {{P}arl{AI}: A dialog
  research software platform}.
\newblock In \emph{Proceedings of the 2017 Conference on Empirical Methods in
  Natural Language Processing: System Demonstrations}, pages 79--84.

\bibitem[{Mir et~al.(2019)Mir, Felbo, Obradovich, and Rahwan}]{mir}
Remi Mir, Bjarke Felbo, Nick Obradovich, and Iyad Rahwan. 2019.
\newblock Evaluating style transfer for text.
\newblock In \emph{Proceedings of the 2019 Conference of the North {A}merican
  Chapter of the Association for Computational Linguistics: Human Language
  Technologies, Volume 1 (Long and Short Papers)}, pages 495--504, Minneapolis,
  Minnesota. Association for Computational Linguistics.

\bibitem[{Montemayor et~al.(2021)Montemayor, Halpern, and
  Fairweather}]{montemayor}
Carlos Montemayor, Jodi Halpern, and Abrol Fairweather. 2021.
\newblock In principle obstacles for empathic ai: why we can’t replace human
  empathy in healthcare.
\newblock \emph{AI \& society}, pages 1--7.

\bibitem[{Mousavi et~al.(2021)Mousavi, Cervone, Danieli, and
  Riccardi}]{mousavi}
Seyed~Mahed Mousavi, Alessandra Cervone, Morena Danieli, and Giuseppe Riccardi.
  2021.
\newblock Would you like to tell me more? generating a corpus of psychotherapy
  dialogues.
\newblock In \emph{Proceedings of the Second Workshop on Natural Language
  Processing for Medical Conversations}, pages 1--9.

\bibitem[{Moyers et~al.(2014)Moyers, Manuel, Ernst, Moyers, Manuel, Ernst, and
  Fortini}]{miti_4_2_1}
TB~Moyers, JK~Manuel, D~Ernst, T~Moyers, J~Manuel, D~Ernst, and C~Fortini.
  2014.
\newblock Motivational interviewing treatment integrity coding manual 4.1 (miti
  4.1).
\newblock \emph{Unpublished manual}.

\bibitem[{Moyers et~al.(2003)Moyers, Martin, Manuel, Miller, and
  Ernst}]{moyers}
Theresa~B Moyers, Tim Martin, Jennifer~K Manuel, William~R Miller, and D~Ernst.
  2003.
\newblock The motivational interviewing treatment integrity (miti) code:
  Version 2.0.
\newblock \emph{Retrieved from Verf{\"u}bar unter: www. casaa. unm. edu [01.03.
  2005]}.

\bibitem[{Nambisan(2011)}]{nambisan2011information}
Priya Nambisan. 2011.
\newblock Information seeking and social support in online health communities:
  impact on patients' perceived empathy.
\newblock \emph{Journal of the American Medical Informatics Association},
  18(3):298--304.

\bibitem[{Papineni et~al.(2002)Papineni, Roukos, Ward, and Zhu}]{bleu}
Kishore Papineni, Salim Roukos, Todd Ward, and Wei-Jing Zhu. 2002.
\newblock \href {https://doi.org/10.3115/1073083.1073135} {{B}leu: a method for
  automatic evaluation of machine translation}.
\newblock In \emph{Proceedings of the 40th Annual Meeting of the Association
  for Computational Linguistics}, pages 311--318, Philadelphia, Pennsylvania,
  USA. Association for Computational Linguistics.

\bibitem[{P{\'e}rez-Rosas et~al.(2018)P{\'e}rez-Rosas, Sun, Li, Wang, Resnicow,
  and Mihalcea}]{perez2018analyzing}
Ver{\'o}nica P{\'e}rez-Rosas, Xuetong Sun, Christy Li, Yuchen Wang, Kenneth
  Resnicow, and Rada Mihalcea. 2018.
\newblock Analyzing the quality of counseling conversations: the tell-tale
  signs of high-quality counseling.
\newblock In \emph{Proceedings of the Eleventh International Conference on
  Language Resources and Evaluation (LREC 2018)}.

\bibitem[{Popovi{\'c}(2015)}]{chrf}
Maja Popovi{\'c}. 2015.
\newblock \href {https://doi.org/10.18653/v1/W15-3049} {chr{F}: character
  n-gram {F}-score for automatic {MT} evaluation}.
\newblock In \emph{Proceedings of the Tenth Workshop on Statistical Machine
  Translation}, pages 392--395, Lisbon, Portugal. Association for Computational
  Linguistics.

\bibitem[{Prabhumoye et~al.(2018)Prabhumoye, Tsvetkov, Salakhutdinov, and
  Black}]{prabhumoye}
Shrimai Prabhumoye, Yulia Tsvetkov, Ruslan Salakhutdinov, and Alan~W Black.
  2018.
\newblock Style transfer through back-translation.
\newblock \emph{arXiv preprint arXiv:1804.09000}.

\bibitem[{Rao and Tetreault(2018)}]{GYAFC}
Sudha Rao and Joel Tetreault. 2018.
\newblock Dear sir or madam, may {I} introduce the {GYAFC} dataset: Corpus,
  benchmarks and metrics for formality style transfer.
\newblock In \emph{Proceedings of the 2018 Conference of the North {A}merican
  Chapter of the Association for Computational Linguistics: Human Language
  Technologies, Volume 1 (Long Papers)}, pages 129--140. Association for
  Computational Linguistics.

\bibitem[{Reimers and Gurevych(2019)}]{sbert}
Nils Reimers and Iryna Gurevych. 2019.
\newblock Sentence-{BERT}: Sentence embeddings using {S}iamese {BERT}-networks.
\newblock In \emph{Proceedings of the 2019 Conference on Empirical Metho0ds in
  Natural Language Processing and the 9th International Joint Conference on
  Natural Language Processing (EMNLP-IJCNLP)}, pages 3982--3992, Hong Kong,
  China. Association for Computational Linguistics.

\bibitem[{Roller et~al.(2021)Roller, Dinan, Goyal, Ju, Williamson, Liu, Xu,
  Ott, Smith, Boureau, and Weston}]{blender}
Stephen Roller, Emily Dinan, Naman Goyal, Da~Ju, Mary Williamson, Yinhan Liu,
  Jing Xu, Myle Ott, Eric~Michael Smith, Y-Lan Boureau, and Jason Weston. 2021.
\newblock Recipes for building an open-domain chatbot.
\newblock In \emph{Proceedings of the 16th Conference of the European Chapter
  of the Association for Computational Linguistics: Main Volume}, pages
  300--325, Online. Association for Computational Linguistics.

\bibitem[{Schwartz(2021)}]{self}
Robert Schwartz. 2021.
\newblock \href
  {https://societyforpsychotherapy.org/the-big-reveal-ethical-implications-of-therapist-self-disclosure}
  {The big reveal | ethical implications of therapist self-disclosure}.

\bibitem[{Shang et~al.(2019)Shang, Li, Fu, Bing, Zhao, Shi, and
  Yan}]{shang2019}
Mingyue Shang, Piji Li, Zhenxin Fu, Lidong Bing, Dongyan Zhao, Shuming Shi, and
  Rui Yan. 2019.
\newblock Semi-supervised text style transfer: Cross projection in latent
  space.
\newblock In \emph{Proceedings of the 2019 Conference on Empirical Methods in
  Natural Language Processing and the 9th International Joint Conference on
  Natural Language Processing (EMNLP-IJCNLP)}, pages 4937--4946. Association
  for Computational Linguistics.

\bibitem[{Sharma et~al.(2020{\natexlab{a}})Sharma, Choudhury, Althoff, and
  Sharma}]{sharma2020engagement}
Ashish Sharma, Monojit Choudhury, Tim Althoff, and Amit Sharma.
  2020{\natexlab{a}}.
\newblock Engagement patterns of peer-to-peer interactions on mental health
  platforms.
\newblock In \emph{Proceedings of the International AAAI Conference on Web and
  Social Media}, volume~14, pages 614--625.

\bibitem[{Sharma et~al.(2020{\natexlab{b}})Sharma, Miner, Atkins, and
  Althoff}]{sharma2020computational}
Ashish Sharma, Adam~S Miner, David~C Atkins, and Tim Althoff.
  2020{\natexlab{b}}.
\newblock A computational approach to understanding empathy expressed in
  text-based mental health support.
\newblock \emph{arXiv preprint arXiv:2009.08441}.

\bibitem[{Sheikha and Inkpen(2011)}]{sheikha2011}
Fadi~Abu Sheikha and Diana Inkpen. 2011.
\newblock Generation of formal and informal sentences.
\newblock In \emph{Proceedings of the 13th European Workshop on Natural
  Language Generation}, pages 187--193.

\bibitem[{Shen et~al.(2017)Shen, Lei, Barzilay, and Jaakkola}]{yelp}
Tianxiao Shen, Tao Lei, Regina Barzilay, and Tommi Jaakkola. 2017.
\newblock Style transfer from non-parallel text by cross-alignment.
\newblock \emph{Advances in neural information processing systems}, 30.

\bibitem[{Steel et~al.(2014)Steel, Marnane, Iranpour, Chey, Jackson, Patel, and
  Silove}]{mental-disorder}
Zachary Steel, Claire Marnane, Changiz Iranpour, Tien Chey, John~W Jackson,
  Vikram Patel, and Derrick Silove. 2014.
\newblock The global prevalence of common mental disorders: a systematic review
  and meta-analysis 1980--2013.
\newblock \emph{International journal of epidemiology}, 43(2):476--493.

\bibitem[{Tatman(2022)}]{tatman}
Rachael Tatman. 2022.
\newblock \href
  {https://makingnoiseandhearingthings.com/2022/08/03/large-language-models-cannot-replace-mental-health-professionals/}
  {[link]}.

\bibitem[{Tian et~al.(2018)Tian, Hu, and Yu}]{pos}
Youzhi Tian, Zhiting Hu, and Zhou Yu. 2018.
\newblock Structured content preservation for unsupervised text style transfer.
\newblock \emph{arXiv preprint arXiv:1810.06526}.

\bibitem[{Wan et~al.(2015)Wan, Jun, Zhang, Pan, and Hua}]{kappa}
TANG Wan, HU~Jun, Hui Zhang, WU~Pan, and HE~Hua. 2015.
\newblock Kappa coefficient: a popular measure of rater agreement.
\newblock \emph{Shanghai archives of psychiatry}, 27(1):62.

\bibitem[{Welivita and Pu(2022)}]{heal}
Anuradha Welivita and Pearl Pu. 2022.
\newblock Heal: A knowledge graph for distress management conversations.

\bibitem[{Xie(2017)}]{dipsy}
Xing Xie. 2017.
\newblock \href
  {https://www.microsoft.com/en-us/research/project/dipsy-digital-psychologist/}
  {Dipsy: A digital psychologist}.

\bibitem[{Xu et~al.(2019)Xu, Ge, and Wei}]{xu2019}
Ruochen Xu, Tao Ge, and Furu Wei. 2019.
\newblock Formality style transfer with hybrid textual annotations.
\newblock \emph{arXiv preprint arXiv:1903.06353}.

\bibitem[{Zhang et~al.(2019)Zhang, Kishore, Wu, Weinberger, and
  Artzi}]{bertscore}
Tianyi Zhang, Varsha Kishore, Felix Wu, Kilian~Q Weinberger, and Yoav Artzi.
  2019.
\newblock Bertscore: Evaluating text generation with bert.
\newblock \emph{arXiv preprint arXiv:1904.09675}.

\end{thebibliography}
\bibliographystyle{acl_natbib}

\appendix

\section{Datasets}

\subsection{The RED (Reddit Emotional Distress) Dataset}
\label{sec:red}

The RED dataset is curated from carefully selected 8 mental health-related subreddits in Reddit. According to the latest statistics, 61\% of Reddit users are male. Of the users, 48\% are from the United States. People aged 18-29 make up Reddit’s largest user base (64\%). The second biggest age group is 30-49 (29\%). Only 7\% of Reddit users are over 50. It should be noted that these demographic biases can subtly skew our data and models from representing average human behavior. The data we curated were English-only and they may perpetuate an English bias in NLP systems. 

%Since the dataset is English-only potentially perpetuates an English bias in NLP systems.

% The fact that the dataset is curated from Reddit can perpetuate certain demographic biases. 

\subsection{The MI Dataset}
\label{sec:mi_labels}

\begin{table*}[ht!]
\small
\centering
\begin{tabularx}{\linewidth}{X p{7cm} p{4.85cm}}
\toprule
\textbf{MITI label} & \textbf{Description} & \textbf{Examples}\\
\midrule

1. Closed Question & Questions that can be answered with an yes/no response or a very restricted range of answers. & \textit{Do you think this is an advantage?} \newline \textit{Did you use herion this week?}\vspace{1mm}\\

2. Open Question & Questions that allow a wide range of possible answers. It may seek information or may invite the speaker’s perspective or may encourage self-exploration.\vspace{1mm} & \textit{What do you think are the advantages of changing this behavior?} \newline \textit{What is your take on that?}\\

3. Simple Reflection & Simple reflections include repetition, rephrasing, or paraphrasing of speaker’s previous statement. It conveys understanding or facilitate speaker-listener exchanges.\vspace{1mm} & \textit{It seems that you are not sure what is going to come out of this talk.} \newline \textit{It sounds like you're feeling worried.}\\

4. Complex Reflection & Complex reflections include repeating or rephrasing the previous statement of the speaker but adding substantial meaning or emphasis to it. It serves the purpose of conveying a deeper or more complex picture of what the speaker has said. & \textbf{Speaker:} \textit{Mostly, I would change for future generations. If we waste everything, then there will be nothing left.} \newline \textbf{Listener:} \textit{It sounds like you have a strong feeling of responsibility.}\vspace{1mm}\\

5. Give Information & The listener gives information, educates, provides feedback, or gives an opinion without advising. & \textit{This assignment on logging your cravings is important because we know that cravings often lead to relapses.}\vspace{1mm}\\

%\midrule
\multicolumn{2}{l}{\textbf{MI Adherent Behaviour Codes:\vspace{1mm}}}\\
%\midrule
6. Advise with Permission & Advising when the speaker asks directly for the information or advice. Indirect forms of permission can also occur, such as when the listener invites the speaker to disregard the advice as appropriate.\vspace{1mm} & \textit{If you agree with it, we could try to brainstorm some ideas that might help you.}\\

7. Affirm & Encouraging the speaker by saying something positive or complimentary. \vspace{1mm} & \textit{You should be proud of yourself for your past’s efforts.}\\

8. Emphasize Autonomy & Emphasizing the speaker’s control, freedom of choice, autonomy, and ability to decide. & \textit{Yes, you’re right. No one can force you to stop drinking.} \newline \textit{It is really up to you to decide.} \vspace{1mm}\\

9. Support & Supporting the client with statements of compassion or sympathy. & \textit{I’m here to help you with this} \newline \textit{I know it’s really hard to stop drinking}\vspace{1mm}\\
\multicolumn{3}{l}{\textbf{MI Non-Adherent Behaviour Codes:\vspace{1mm}}}\\

10. Advise without Permission & Making suggestions, offering solutions or possible actions without first obtaining permission from the speaker.  & \textit{You should simply scribble a note that reminds you to turn the computer off during breaks.}\vspace{1mm}\\

11. Confront & Directly and unambiguously disagreeing, arguing, correcting, shaming, blaming, criticizing, labeling, moralizing, ridiculing, or questioning the speaker’s honesty.  & \textit{You think that is any way to treat people you love?} \newline \textit{Yes, you are an alcoholic. You might not think so, but you are.}\vspace{1mm}\\

12. Direct & Giving the speaker orders, commands, or imperatives. & \textit{Don’t do that!} \newline \textit{Keep track of your cravings, using this log, and bring it in next week to review with me.}\vspace{1mm}\\

13. Warn & A statement or event that warns of something or that serves as a cautionary example. & \textit{Be careful, DO NOT stop taking meds without discussing with your doctor.} \vspace{1mm}\\

\multicolumn{3}{l}{\textbf{Other:\vspace{1mm}}}\\

14. Self-Disclose & The listener discloses his/her personal information or experiences. & \textit{I used to be similar where I get obsessed about how people look but after maturing some I got over that. }\vspace{1mm}\\

15. Other & All other statements that are not classified under any of the above codes & \textit{Good morning.} \newline \textit{Hi there.}\\
\bottomrule
\end{tabularx}
\caption{The set of labels adapted from the MITI code that the MI classifier is able to recognize.}
\label{table:miti}
\end{table*}

Altogether, 15 labels adapted from the MITI code 2.0 \cite{moyers} and 4.2.1 \cite{miti_4_2_1} were used for annotation. They included \textit{Closed Question}, \textit{Open Question}, \textit{Simple Reflection}, \textit{Complex Reflection}, and \textit{Give Information}, which are generally considered favourable. They also included labels recognized specifically as MI adherent, which are \textit{Advise with Permission}, \textit{Affirm}, \textit{Emphasize Autonomy}, and \textit{Support}. There are another four labels recognized as MI non-adherent, which are \textit{Advise without Permission}, \textit{Confront}, \textit{Direct}, and \textit{Warn}. We also included two other labels \textit{Self-Disclose} and \textit{Other}, which are not included in the MITI code. The label \textit{Self-Disclose} was included because, in peer support conversations, peers are mostly seen to share their lived experiences. Though it is believed that \textit{Self-Disclosure} contributes in building rapport between the speaker and listener, as suggested by R. Schwartz \shortcite{self}, this type of disclosure must be used wisely with caution since it can as well be counterproductive distorting client’s transference. Thus, it is important to be able to recognize this response type. 

Table \ref{table:miti} shows the full list of labels we adapted from the MITI code along with descriptions and examples. Table \ref{table:dist_stats} shows the statistics of the annotated responses in the MI dataset, corresponding to each label. 

\begin{table}[ht!]
\small
\centering
%\begin{tabularx}{\linewidth}{X p{1cm} p{1cm} p{1cm}}
\begin{tabularx}{\linewidth}{X | r r | r}
\toprule

\textbf{Label} & \textbf{\# Labels} & \textbf{\# Labels} & \textbf{Total}\\

 & \textbf{in CC} & \textbf{in RED}\\
 
%  &  &  & \textbf{compared to RED} & \textbf{to CounselChat}\\

\midrule

Closed Question & 500 & 405 & 905\\

Open Question & 264 & 212 & 476\\

Simple Reflection & 304 & 252 & 556\\

Complex Reflection & 732 & 562 & 1,294\\

Give Information & 3,643 & 1213 & 4,856\vspace{1mm}\\

\multicolumn{3}{l}{\textbf{MI Adherent Behavior Codes:}\vspace{1mm}}\\

Advise w/ Permission & 417 & 67 & 484\\

Affirm & 428 & 517 & 945\\

Emphasize Autonomy & 152 & 101 & 253\\

Support & 418 & 815 & 1,233\vspace{1mm}\\

\multicolumn{3}{l}{\textbf{MI Non-Adherent Behavior Codes:}\vspace{1mm}}\\

Advise w/o Permission & 1,414 & 871 & 2,285\\

Confront & 142 & 176 & 318\\

Direct & 460 & 438 & 898\\

Warn & 67 & 46 & 113\vspace{1mm}\\

\multicolumn{3}{l}{\textbf{Other:}\vspace{1mm}}\\

Self-Disclose & 174 & 1216 & 1,390\\

Other & 513 & 292 & 805\\

\midrule

Total & 9,628 & 7,183 & 16,811\\

\bottomrule
\end{tabularx}
\caption{Statistics of human annotated MITI labels in CounselChat (CC) and RED datasets.}
\label{table:dist_stats}
\end{table}

\subsection{Data Augmentation: N-gram Based Matching}
\label{sec:ngram}

We denote examples of the most frequent N-grams corresponding to each label in Table \ref{table:ngrams}. For simplicity, we list only some of them along with their corresponding frequencies. For data augmentation, we used all four-grams and five-grams, which had a frequency of above 5.    

\begin{table*}[ht!]
\small
\centering
%\begin{tabularx}{\linewidth}{X p{1cm} p{1cm} p{1cm}}
\begin{tabularx}{\linewidth}{l | X | X}
\toprule

\textbf{Label} & \textbf{Examples of most frequent four-grams} & \textbf{Examples of most frequent five-grams} \\

\midrule

Closed Question & \textit{Do you have any} (11), \textit{Do you have a} (7), \textit{Do you want to} (7), \textit{Have you talked to} (5), \textit{Do you think you} (5) & - \\

Open Question & \textit{Do you want to} (10), \textit{you want to be} (8), \textit{How do you feel} (5), \textit{Why do you feel} (5), \textit{What is the evidence} (5) & \textit{Do you want to be} (6)\\

Simple Reflection & \textit{It sounds like you} (16), \textit{sounds like you have} (9), \textit{sounds like you are} (8) & \textit{It sounds like you are} (7), \textit{It sounds like you have} (6)\\

Complex Reflection & \textit{It sounds like you} (26), \textit{My guess is that} (5), \textit{The fact that you} (5), \textit{why you might feel} (5) & \textit{It sounds like you are} (7), \textit{It sounds like you have} (6)\\

Give Information & \textit{may be able to} (11), \textit{who you are and} (8), \textit{For example , if} (8), \textit{A lot of people} (7), \textit{A good therapist will} (6) & \textit{who you are and what} (6), \textit{you are and what you} (6), \textit{be able to help you} (6), \textit{it is important to} (5), \textit{a higher level of care} (5)\vspace{1mm}\\

%\multicolumn{3}{l}{\textbf{MI Adherent Behavior Codes:}\vspace{1mm}}\\

Advise w/ Permission & \textit{It may be helpful} (8), \textit{would be a good} (7), \textit{you would like to} (6), \textit{a good idea to}	(5), \textit{I would encourage you} (5) & \textit{It may be helpful to} (6), \textit{I would encourage you to} (5)\\

Affirm & \textit{I 'm glad you} (19), \textit{wish you the best} (7), \textit{I 'm glad that} (7), \textit{I wish you the} (6), \textit{you 're doing better} (5) & \textit{I 'm glad you 're} (9), \textit{I wish you the best} (6)\\

Emphasize Autonomy & - & -\\

Support & \textit{I 'm so sorry} (12), \textit{sorry to hear about} (12), \textit{I hope you find} (10), \textit{you are not alone} (9), \textit{m here for you} (8) & \textit{I 'm sorry to hear} (11), \textit{I 'm here for you} (8), \textit{I know how you feel} (8), \textit{if you wan na talk} (6), \textit{I hope you can find}	(5)\vspace{1mm}\\

%\multicolumn{3}{l}{\textbf{MI Non-Adherent Behavior Codes:}\vspace{1mm}}\\

Advise w/o Permission & \textit{Reach out to a} (6), \textit{I would suggest that} (6), \textit{I think you should} (5), \textit{I urge you to} (5), \textit{I think you need} (5) & \textit{, you may want to} (5), \textit{I would suggest that you} (5)\\

Confront & - & -\\

Direct & - & -\\

Warn & - & -\vspace{1mm}\\

%\multicolumn{3}{l}{\textbf{Other:}\vspace{1mm}}\\

Self-Disclose & \textit{I feel the same} (9), \textit{I 've been in} (8), \textit{the same way .}	(7),  \textit{do n't know what} (6), \textit{I feel like it} (5) & \textit{I feel the same way} (5), \textit{I do n't know what}	(5)\\

Other & \textit{you for your question} (12), \textit{Hello , and thank} (9), \textit{thank you for your} (9) & \textit{Hello , and thank you} (9), \textit{you for your question .} (12)\\

\bottomrule
\end{tabularx}
\caption{Examples of most frequent four-grams and five-grams corresponding to each label. Their frequencies are denoted within brackets.}
\label{table:ngrams}
\end{table*}

Table \ref{table:ngramretrieve} shows the statistics of the labels extended through N-gram based matching in CC and RED datasets. We also encountered 518 and 53,196 sentences in CounselChat and RED datasets respectively that had overlapping labels, which were discarded due to ambiguity.

\subsection{Data Augmentation: Similarity Based Retrieval}
\label{sec:retrieval}

% When retrieving labeled sentences corresponding to each unlabeled sentence, 

To derive semantically meaningful sentence embeddings that can be compared using cosine-similarity, we used Sentence-BERT (SBERT) proposed by Reimers and Gurevych \shortcite{sbert}, which uses siamese and triplet network structures to compute sentence embeddings. Among several models the authors have proposed, we used the \textit{roberta-base-nli-stsb-mean-tokens} model, fine-tuned on the NLI \cite{A_large_annotated_corpus_for_learning_natural_language_inference} and STS benchmark (STSb) \cite{SemEval-2017_Task1_Semantic_Textual_Similarity_Multilingual_and_Crosslingual_Focused_Evaluation} datasets, since it has reported a high Spearman's rank correlation of  ${84.79 \pm 0.38}$  between the cosine-similarity of the sentence embeddings and the gold labels in the STS benchmark test set outperforming the existing state-of-the-art. It is also more efficient to use than \textit{roberta-large}.

\begin{figure}[ht!]
  \centering
  \includegraphics[width=\linewidth]{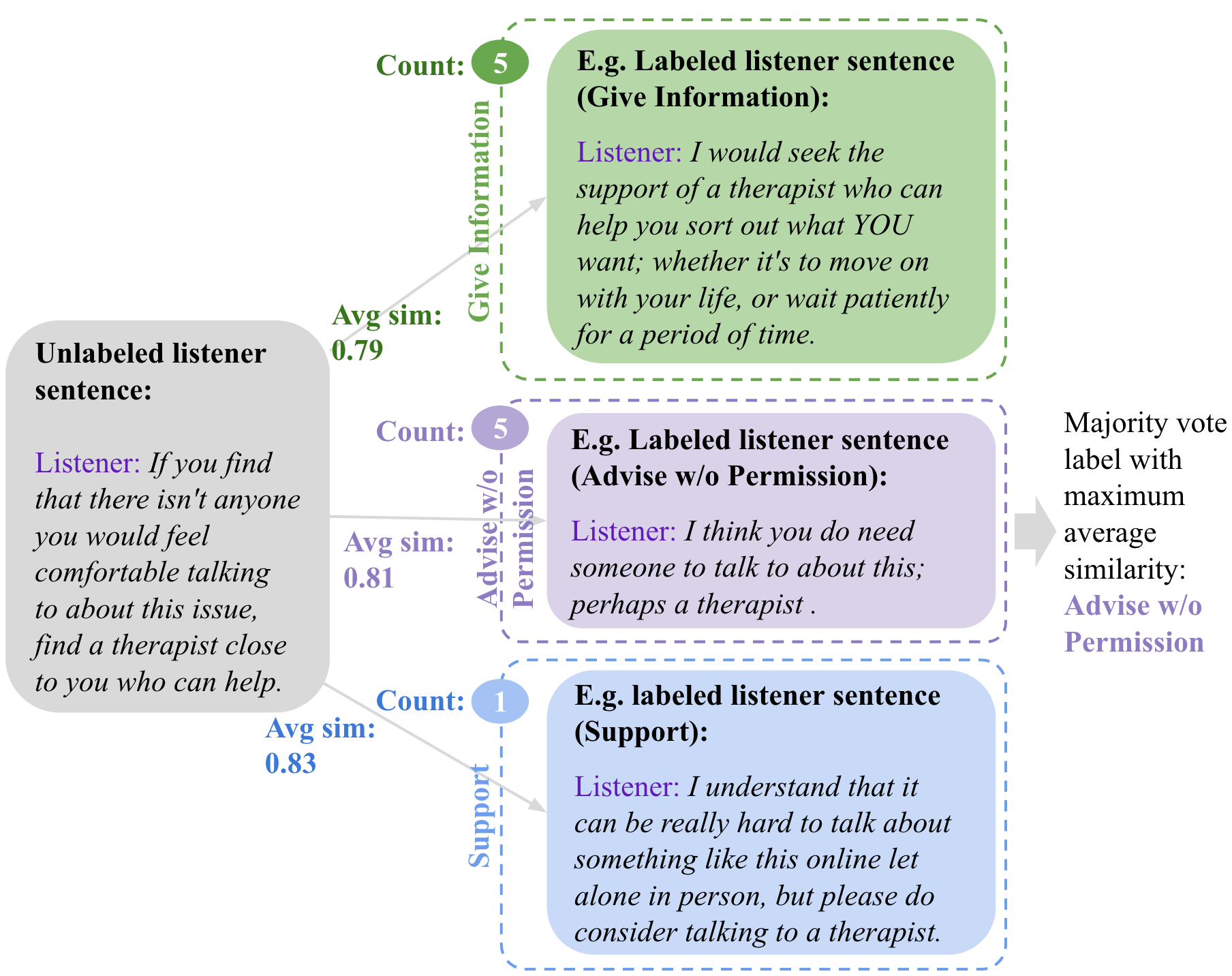}
  \caption{An example of automatically labeling an unlabeled sentence by computing the cosine-similarity with labeled sentences. The label is chosen based on majority voting. But this example shows a tie. Thus, we compute the average similarity of the sentence clusters that hold a tie and select the label of the sentence cluster with the maximum average similarity.}
  \label{fig:sim_ex}
\end{figure}

As described in Section \ref{sec:datasets}, we used majority voting followed by computing the average similarity of retrieved sentences with the same label (in case of ties) to choose the final label for an unlabeled sentence. In Figure \ref{fig:sim_ex}, we show an example elaborating this procedure.  

% To compute the sentence embeddings, we used Sentence-BERT (SBERT) proposed by Reimers and Gurevych \shortcite{sbert}, which uses siamese and triplet network structures to derive semantically meaningful sentence embeddings that can be compared using cosine-similarity. 

%Figure \ref{fig:sim_ex} shows an example scenario. 

%To do: Include statistics of the data augmented with similarity based retrieval.

Table \ref{table:ngramretrieve} shows the statistics of the labels extended through similarity-based retrieval in CC and RED datasets.

\begin{table*}[ht!]
\small
\centering
%\begin{tabularx}{\linewidth}{X p{1cm} p{1cm} p{1cm}}
%\begin{tabularx}{\linewidth}{l | r r r | r r r}
\begin{tabular}{l | r r r | r r r}
\toprule

\textbf{Label} & \multicolumn{3}{c}{\textbf{N-gram based matching}} & \multicolumn{3}{c}{\textbf{Similarity-based retrieval}}\vspace{1mm} \\

 & \textbf{\# Labels} & \textbf{\# Labels} & \textbf{Total} & \textbf{\# Labels} & \textbf{\# Labels} & \textbf{Total}\\

 & \textbf{in CC} & \textbf{in RED} & & \textbf{in CC} & \textbf{in RED} & \\

\midrule

Closed Question & 75 & 17,190 & 17,265 & 132 & 71,505 & 61,637 \\

Open Question & 29 & 12,242 & 12,271 & 49 & 36,107 & 36,156 \\

Simple Reflection & 71 & 9,674 & 9,745 & 43 & 21,827 & 21,870 \\

Complex Reflection & 110 & 20,539 & 20,649 & 20 & 17,243 & 17,263 \\

Give Information & 571 & 71,996 & 72,567 & 893 & 166,586 & 167,479 \vspace{1mm}\\

%\multicolumn{3}{l}{\textbf{MI Adherent Behavior Codes:}\vspace{1mm}}\\

Advise w/ Permission & 161 & 5,979 & 6,140 & 5 & 3,728 & 3,733 \\

Affirm & 136 & 16,407 & 16,543 & 187 & 106,066 & 106,253 \\

Emphasize Autonomy & 0 & 0 & 0 & 3 & 2,839 & 2,842 \\

Support & 213 & 94,670 & 94,883 & 482 & 528,469 & 528,951 \vspace{1mm}\\

%\multicolumn{3}{l}{\textbf{MI Non-Adherent Behavior Codes:}\vspace{1mm}}\\

Advise w/o Permission & 520 & 58,857 & 59,377 & 969 & 171,502 & 172,471 \\

Confront & 0 & 0 & 0 & 1 & 2,581 & 2,582\\

Direct & 0 & 0 & 0 & 16 & 21,058 & 21,074\\

Warn & 0 & 0 & 0 & 6 & 2,342 & 2,348\vspace{1mm}\\

%\multicolumn{3}{l}{\textbf{Other:}\vspace{1mm}}\\

Self-Disclose & 5 & 28,309 & 28,314 & 8 & 14,702 & 14,710\\

Other & 27 & 4,498 & 4,525 & 67 & 29,457 & 28,524\\

\midrule

Total & 1,918 & 340,361 & 342,279 & 2,881 & 1,196,012 & 1,198,893 \\

\bottomrule
\end{tabular}
\caption{Statistics of the labels extended through N-
gram-based matching and similarity-based retrieval in CC and RED datsets.}
\label{table:ngramretrieve}
\end{table*}

\subsection{Augmented MI Datasets}
\label{sec:unionintersection}

Table \ref{table:unionintersection} shows the statistics corresponding to each label in the MI Augmented (Union) and MI Augmented (Intersection) datasets developed by taking the union and the intersection of the sentences automatically annotated by N-gram based matching and similarity based retrieval methods.

\begin{table*}[ht!]
\small
\centering
%\begin{tabularx}{\linewidth}{X p{1cm} p{1cm} p{1cm}}
%\begin{tabularx}{\linewidth}{l | r r r | r r r}
\begin{tabular}{l | r r r r | r r r r}
\toprule

\textbf{Label} & \multicolumn{4}{c}{\textbf{MI Augmented (Intersection)}} & \multicolumn{4}{c}{\textbf{MI Augmented (Union)}}\vspace{1mm} \\

 & \textbf{\# Labels} & \textbf{\# Labels} & \textbf{Total} & \textbf{Total} & \textbf{\# Labels} & \textbf{\# Labels} & \textbf{Total} & \textbf{Total}\\

 & \textbf{in CC} & \textbf{in RED} & & \textbf{+ MI Gold} & \textbf{in CC} & \textbf{in RED} & & \textbf{+ MI Gold} \\

\midrule

Closed Question & 9 & 5,598 & 5,607 & \textbf{6,512} & 135 & 78,932 & 79,067 & \textbf{79,972} \\ 
%905

Open Question & 1 & 2,353 & 2,354 & \textbf{2,830} & 60 & 40,805 & 40,865 & \textbf{41,341} \\ 
%476

Simple Reflection & 1 & 185 & 186 & \textbf{742} & 41 & 19,961 & 20,002 & \textbf{20,558} \\ 
%556

Complex Reflection & 2 & 201 & 203 & \textbf{1,497} & 44 & 21,247 & 21,291 & \textbf{22,585} \\ 
%1,294

Give Information & 77 & 3,379 & 3,456 & \textbf{8,312} & 1083 & 203,110 & 204,193 & \textbf{209,049} \\ %4,856

%\multicolumn{3}{l}{\textbf{MI Adherent Behavior Codes:}\vspace{1mm}}\\

Advise w/ Permission & 0 & 28 & 28 & \textbf{512} & 5 & 3,052 & 3,057 & \textbf{3,541} \\ 
%484

Affirm & 48 & 898 & 946 & \textbf{1,891} & 208 & 106,575 & 106,783 & \textbf{107,728} \\ 
%945

Emphasize Autonomy & 0 & 0 & 0 & \textbf{253} & 3 & 2,700 & 2,703 & \textbf{2,956} \\ 
%253

Support & 76 & 44,635 & 44,711 & \textbf{45,944} & 551 & 592,220 & 592,771 & \textbf{594,004} \\ 
%1,233

%\multicolumn{3}{l}{\textbf{MI Non-Adherent Behavior Codes:}\vspace{1mm}}\\

Advise w/o Permission & 144 & 8,872 & 9,016 & \textbf{11,301} & 1,029 & 196,571 & 197,600 & \textbf{199,885} \\ 
%2,285

Confront & 0 & 0 & 0 & \textbf{318} & 0 & 2,468 & 2,468 & \textbf{2,786} \\ 
%318

Direct & 0 & 0 & 0 & \textbf{898} & 15 & 20,690 & 20,705 & \textbf{21,603} \\ 
%898

Warn & 0 & 0 & 0 & \textbf{113} & 6 & 2,278 & 2,284 & \textbf{2,397} \\ 
%113

%\multicolumn{3}{l}{\textbf{Other:}\vspace{1mm}}\\

Self-Disclose & 0 & 729 & 729 & \textbf{2,119} & 12 & 36,522 & 36,534 & \textbf{37,924} \\ 
%1,390

Other & 0 & 5 & 5 & \textbf{810} & 67 & 31,268 & 31,335 & \textbf{32,140} \\ 
%805

\midrule

Total & 358 & 66,883 & 67,241 & \textbf{84,052} & 3,259 & 1,358,399 & 1,361,658 & \textbf{1,378,469}\\ 

% 1,305,464

%Total & 358 & 66,883 & 67,241 & \textbf{84,052} & 3,259 & 1,358,399 & 1,361,658 & \textbf{1,378,469}\\ 
%16,811

\bottomrule
\end{tabular}
\caption{Statistics of the annotated responses in MI Augmented (Intersection) and MI Augmented (Union) datasets.}
\label{table:unionintersection}
\end{table*}

\section{MI Classifier}
\label{sec:classifier}

We used the same hyper-parameter setting used in RoBERTa \cite{roberta} when training the MI classifier. We used the Adam optimizer with $\beta_1$ of $0.9$, $\beta_2$ of $0.98$, an $\epsilon$ value of $1\times10^{-6}$, and a learning rate of $2\times10^{-5}$. A dropout of 0.1 was used on all layers and attention weights, and a GELU activation function \cite{hendrycks2016gaussian}. We limited the maximum number of input tokens to $100$, and used a batch size of $32$. All models were trained for 20 epochs. In all cases, the optimal epoch was selected based on the average cross entropy loss calculated between the ground-truth and predicted labels of the human-annotated (MI Gold) validation set. All the experiments were conducted on a machine with 2x12cores@2.5GHz, 256 GB RAM, 2x200 GB SSD, and 4xGPU (NVIDIA Titan X Pascal). Experiments were also done using GPT3 as the pre-trained language model, however, RoBERTa was seen to outperform GPT3 in this classification task.

\begin{figure}[ht!]
  \centering
  \includegraphics[width=\linewidth]{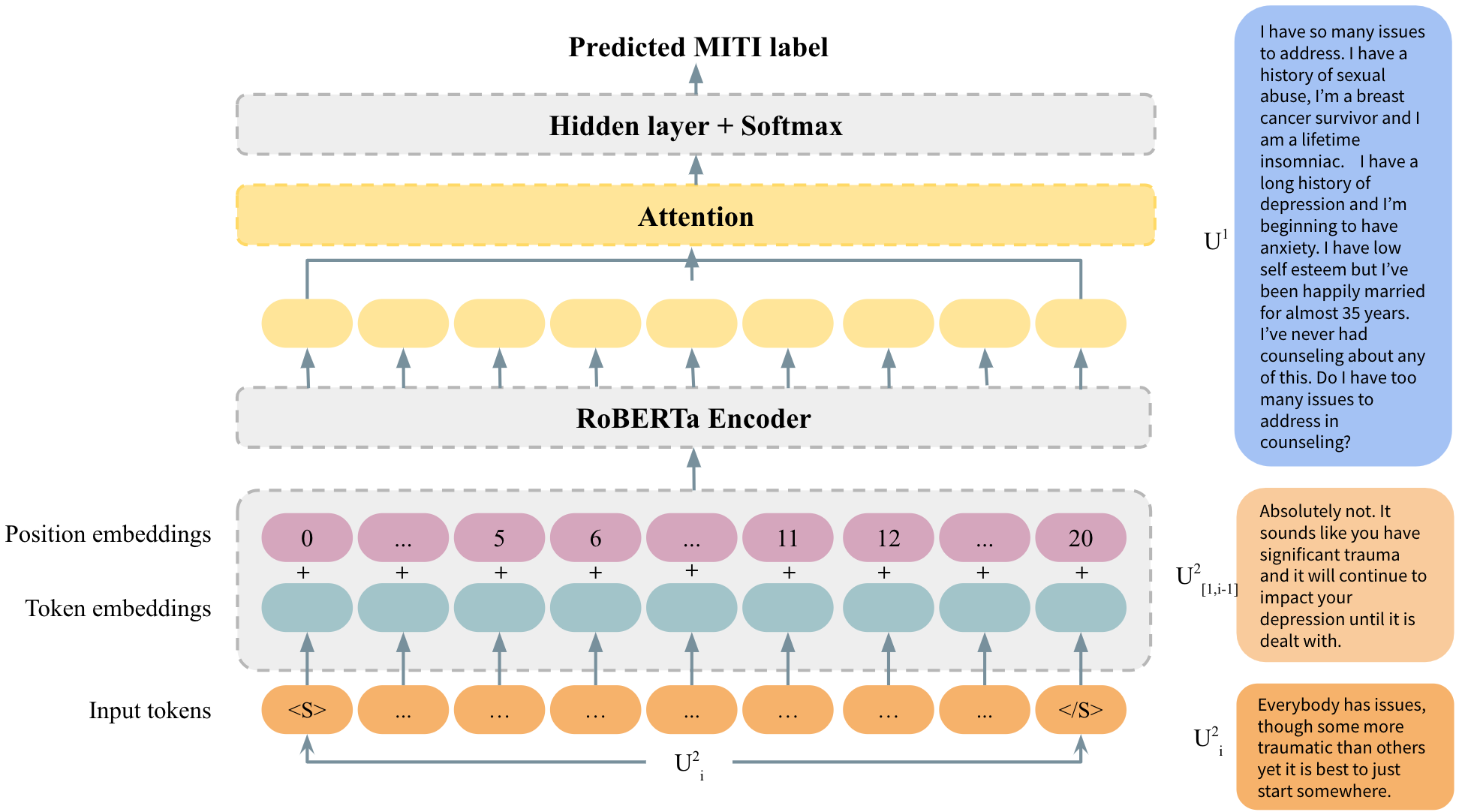}
  \caption{The architecture of the MI classifier.}
  \label{fig:classifier}
\end{figure}

Figure \ref{fig:classifier} shows the architectural diagram of the MI classifier used for annotation. Table \ref{table:classifier} shows the performance scores of the MI classifier when trained on gold-labeled and augmented MI datasets.

%We trained three classifiers. The first294 one was trained on the smaller human-annotated295 MI dataset (MI Gold) taking 80% of the data for296 training and leaving 10% each for validation and297 testing. The other two were trained on the MI Aug-298 mented (Union) and MI Augmented (Intersection)299 datasets, leaving out the data used for validation300 and testing in the first case. In all cases, the optimal301 model was chosen based on the accuracy reported302 on the human-annotated validation set.

%figure and hyperparameter settings, the hardware used to train the models.

\begin{figure}
     \centering
     \subfloat[][Pseudo-Parallel (PP) Corpus]{\includegraphics[width=\linewidth]{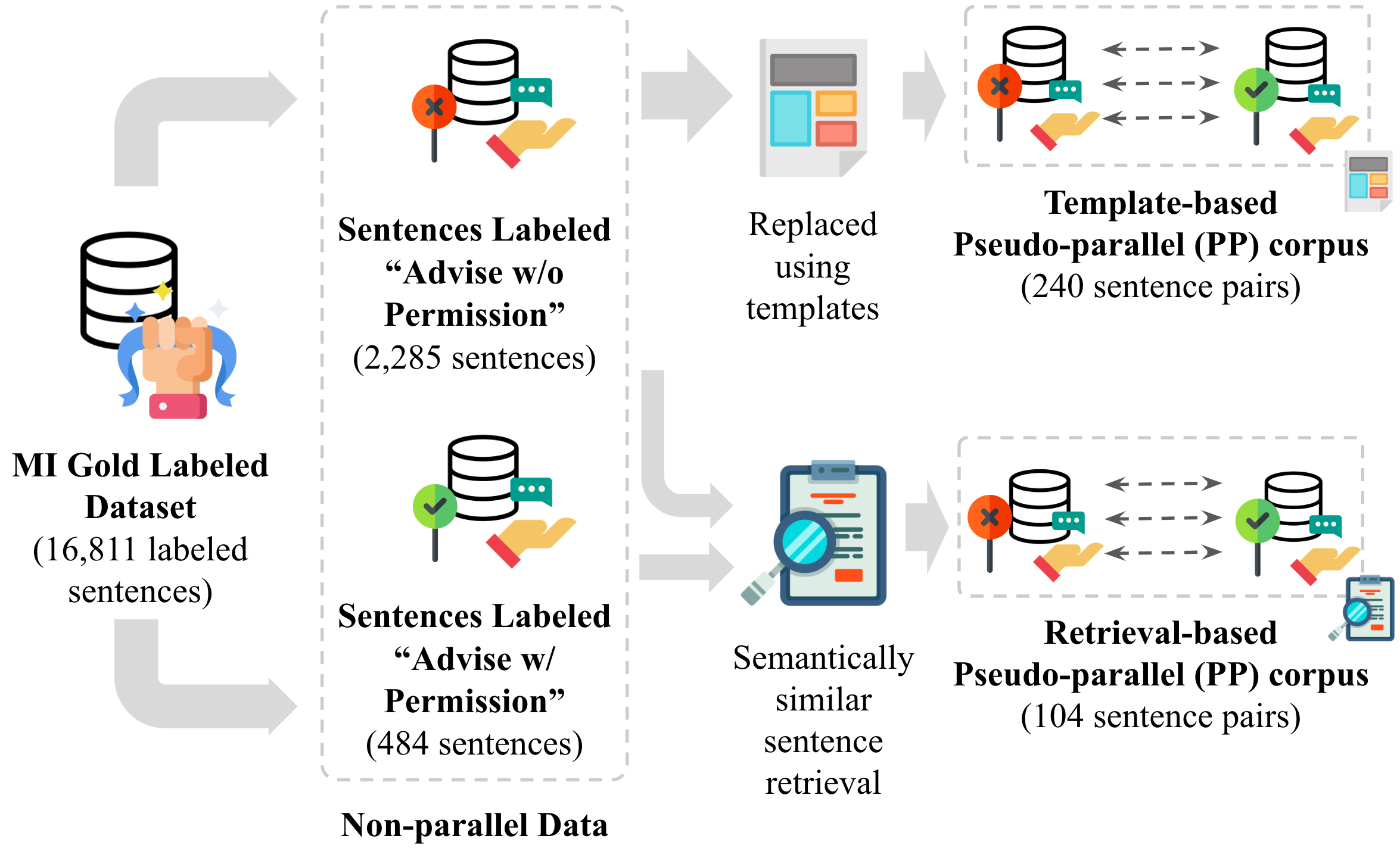}\label{fig:pseudo}}
     \qquad
     \subfloat[][Pseudo-Parallel Augmented (PPA) Corpus]{\includegraphics[width=\linewidth]{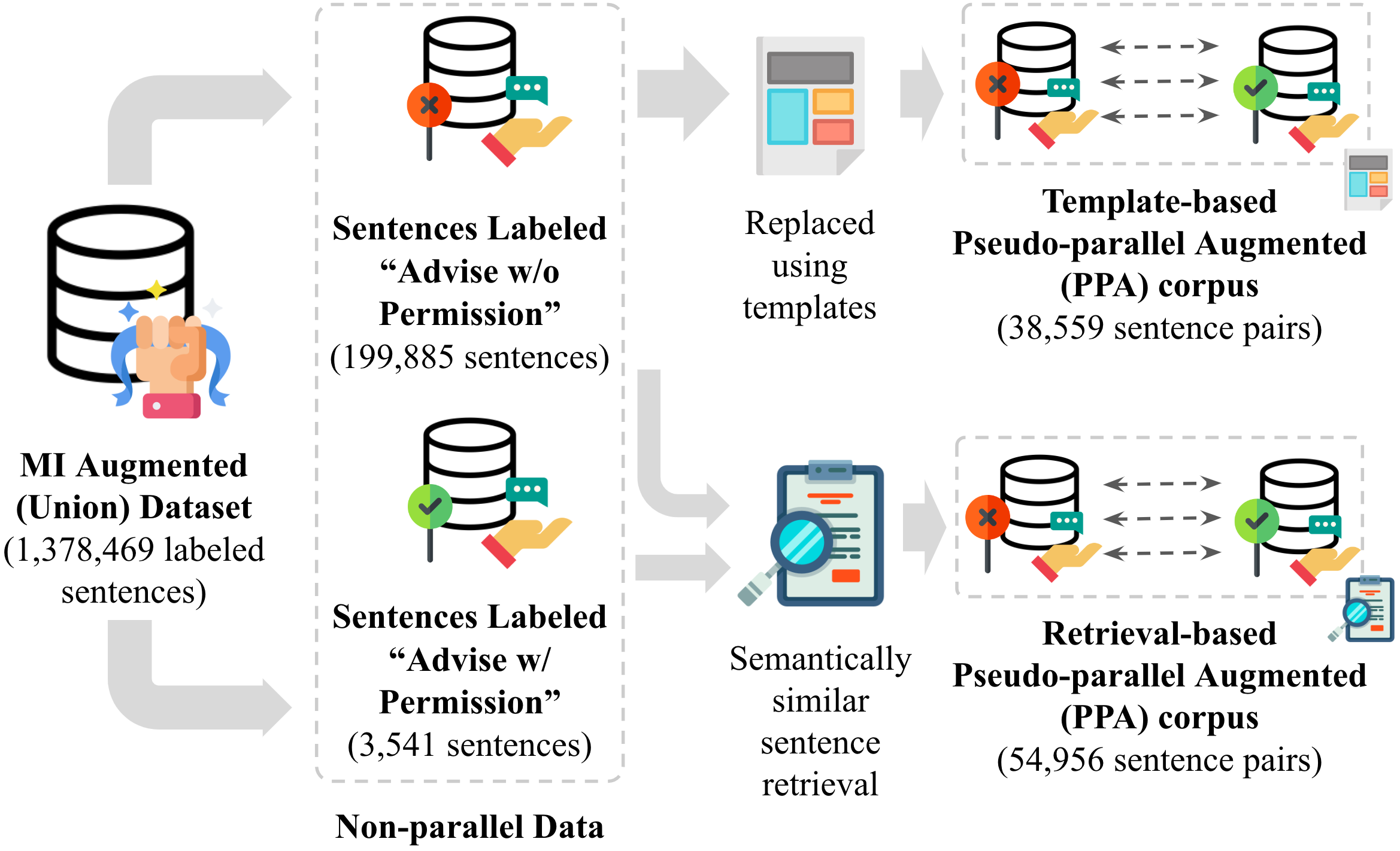}\label{fig:pseudo-augmented}}
     \caption{Pseudo-Parallel (PP) and Pseudo-Parallel Augmented (PPA) corpus construction.}
     %\label{fig-2}
\label{fig:parallel-corpus}
\end{figure}

% According to the results, the labels collected through the union of the N-gram matching and cosine similarity-based methods, which accounts for $\approx$1.3M labels improved the accuracy of the classifier by 8.33\% and 7.5\%, respectively on the validation and test sets compared to the accuracies reported when trained on the gold-labeled MI dataset. 

%up to 72.67\% in the validation set and up to 73.44\% in the test set. This is an increase by 8.33\% and 7.5\%, respectively on the validation and test sets compared to the accuracies when trained on the gold-labeled MI dataset. 

%The confusion matrices associated with each of these classifiers are denoted in the appendices. 

\begin{table*}[ht!]
\small
\centering
%\begin{tabularx}{\linewidth}{X p{1cm} p{1cm} p{1cm}}
%\begin{tabularx}{\linewidth}{X | r r | r r | r r}
\begin{tabular}{l | r r | r r | r | r r}
\toprule

\textbf{Dataset} &  \multicolumn{2}{c|}{\textbf{Size}} & \textbf{Optimal} &  \textbf{Train} & \textbf{Valid} & \multicolumn{2}{c}{\textbf{Test}}\\

 & & & \textbf{Epoch} & \textbf{Loss} &   \textbf{Acc. (\%)} & \textbf{Acc. (\%)} & \textbf{F1-score}\\

 & & & & &  & & \textbf{(weighted avg.)}\\

\midrule

\multirow{3}{*}{MI Gold} & Train: & 13,449 & \multirow{3}{*}{7} & \multirow{3}{*}{0.3002} & \multirow{3}{*}{67.08} & \multirow{3}{*}{68.31} & \multirow{3}{*}{68.07} \\ 
 & Valid (Gold): & 1,681 & & & & \\
 & Test (Gold): & 1,681 & & & & \\

\midrule

 MI   & Train: & 80,690 & \multirow{3}{*}{2} & \multirow{3}{*}{0.2277} & \multirow{3}{*}{64.07} & \multirow{3}{*}{67.13} & \multirow{3}{*}{65.85}\\ 
 Augmented & Valid (Gold): & 1,681 & & & & \\
 (Intersection) & Test (Gold): & 1,681 & & & & \\

 % 80,690

 \midrule

  MI   & Train: & 1,375,107 & \multirow{3}{*}{13} & \multirow{3}{*}{0.1324} & \multirow{3}{*}{\textbf{72.67}} & \multirow{3}{*}{\textbf{73.44}} & \multirow{3}{*}{\textbf{72.92}} \\ 
  % 1,318,913
 Augmented & Valid (Gold): & 1,681 & & & & \\ 
(Union) & Test (Gold): & 1,681 & & & & \\

% 1,318,913

%\multicolumn{3}{l}{\textbf{MI Adherent Behavior Codes:}\vspace{1mm}}\\

\bottomrule
\end{tabular}
\caption{The performance scores of the MI classifier when trained on gold-labeled and augmented MI datasets. All scores are reported on the human-annotated validation and test sets. All scores are reported for a single run.}
\label{table:classifier}
\end{table*}

\section{MI Rephraser}

\subsection{Construction of pseudo-parallel corpora}
\label{sec:pp}

Table \ref{tab:template} denotes the full list of templates corresponding to \textit{Advise without Permission} and \textit{Advise with Permission} responses that were used in the process of creating pseudo-parallel corpora using the template-based replacement method. 

\begin{table*}[ht!]
\small
\centering
%\begin{tabularx}{\linewidth}{X p{1cm} p{1cm} p{1cm}}
%\begin{tabularx}{\linewidth}{p{3cm} | X}
\begin{tabular}{l | l}
\toprule

\textbf{Advise without Permission} &  \textbf{Advise with Permission}\\

\midrule

- \textit{You can} (verb) \underline{\hspace{0.5cm}} & - \textit{It maybe helpful to} (verb) \underline{\hspace{0.5cm}} \\
- \textit{You could} (verb) \underline{\hspace{0.5cm}} & - \textit{You may want to} (verb) \underline{\hspace{0.5cm}}\\
- \textit{You need to} (verb) \underline{\hspace{0.5cm}} & - \textit{I encourage you to} (verb) \underline{\hspace{0.5cm}}\\
- \textit{You should} (verb) \underline{\hspace{0.5cm}} & - \textit{Perhaps you can} (verb) \underline{\hspace{0.5cm}}\\
- (Verb) \underline{\hspace{0.5cm}} & - \underline{\hspace{0.5cm}}\textit{, if you would like.}\\

- \textit{You can try to} (verb) \underline{\hspace{0.5cm}} & - \textit{It would be good idea to} (verb) \underline{\hspace{0.5cm}}\\
- \textit{I think you should} (verb) \underline{\hspace{0.5cm}} & - \textit{It may be important to} (verb) \underline{\hspace{0.5cm}}\\
- \textit{I suggest that you} (verb) \underline{\hspace{0.5cm}} & - \textit{I would encourage you to} (verb) \underline{\hspace{0.5cm}}\\
- \textit{I suggest you} (verb) \underline{\hspace{0.5cm}} & - \textit{I wonder if you can} (verb) \underline{\hspace{0.5cm}}\\
- \textit{Maybe you can} (verb) \underline{\hspace{0.5cm}} & - \textit{Maybe it is important to} (verb) \underline{\hspace{0.5cm}}\\
- \textit{Maybe you could} (verb) \underline{\hspace{0.5cm}} & - \textit{An option would be to} (verb) \underline{\hspace{0.5cm}}\\

     & - \textit{You may want to consider} (present continuous form of the verb) \underline{\hspace{0.5cm}}\\
      & - \textit{You may consider} (present continuous form of the verb) \underline{\hspace{0.5cm}}\\
       & - \textit{I would recommend} (present continuous form of the verb) \underline{\hspace{0.5cm}}\\
  & - \textit{I wonder if you can consider} (present continuous form the verb) \underline{\hspace{0.5cm}}\\

\bottomrule
\end{tabular}
%\end{tabularx}
\caption{Linguistic templates corresponding to \textit{Advise without Permission} and \textit{Advise with Permission} responses.}
\label{tab:template}
\end{table*}

In Figure \ref{fig:parallel-corpus}, we visualize the process of creating Pseudo-Parallel (PP) and Pseudo-Parallel Augmented (PPA) corpora along with statistics corresponding to each dataset.

\subsection{Rephrasing Models}
\label{sec:models}

For developing rephrasing models, we used the 90M parameter version of Blender \cite{blender}. It contains an 8 layer encoder, an 8-layer decoder with 512-dimensional embeddings, and 16 attention heads. It has a maximum input length of 1024 tokens. All code for fine-tuning is available in ParlAI \cite{parlai}. All the models were fine-tuned for 200 epochs, with a batch size of 8, and a learning rate of $1 \times 10\textsuperscript{-6}$. For other hyperparameters, we used the default values defined in their documentation at \url{https://parl.ai/projects/recipes}. Fine-tuning the models was conducted in a machine with 2x12cores@2.5GHz, 256 GB RAM, 2x200 GB SSD, and 4xGPU (NVIDIA Titan X Pascal). 
%1e-06
% We used the default set of hyperparameters defined for fine-tuning Blender at \url{https://parl.ai/projects/recipes}. 
% A learning rate of 1e-06 was used to train Blender based rephraser models on pseudo-parallel corpora developed using different automatic methods.

We also used GPT3 pretrained language model having 175 billion parameters. The smallest but fastest version of GPT3, Ada was used in our experiments. Fine-tuning of GPT3 models were done through the paid API provided by OpenAI (\url{www.openai.com}) following API guide at \url{https://beta.openai.com/docs/guides/fine-tuning}. We used the default set of hyperparameters for fine-tuning all GPT3 based models. These hyperparameters are tested to work well across a range of use cases. All the models were fine-tuned for 4 epochs, with a batch size $\approx$0.2\% of the number of examples in the training set (capped at 256), and a learning rate of 0.05.

Table \ref{tab:examples} shows some examples of rephrased sentences by the different rephraser models we fine-tuned. 

\begin{table*}[ht!]
\small
\centering
%\begin{tabularx}{\linewidth}{X p{1cm} p{1cm} p{1cm}}
\begin{tabularx}{\linewidth}{X}
%\begin{tabular}{l}
\toprule

\textbf{Advise without permission:} \textit{try to learn from your mistakes and meet some new people .} \\

\textbf{Advise with permission (ground-truth):} \textit{It may be important to try to learn from your mistakes and meet some new people .}\vspace{1mm}\\

\textbf{\underline{Template-based:}}\\

\textbf{Blender:} \textit{it would be a good idea to try to learned from your mistakes and meet some new people like me .}\\
\textbf{GPT3:} \textit{I encourage you to try to learn from your mistakes and meet some new people .}\vspace{1mm}\\

\textbf{\underline{Retrieval-based:}}\\

\textbf{Blender:} \textit{In addition to being open to meeting new friends , involve yourself in projects from which you ' ll feel satisfied by your own results .}\\
\textbf{GPT3:} \textit{I think it would be valuable for you to start mingling with other young people .}\vspace{1mm}\\

\textbf{\underline{Template+Retrieval-based:}}\\

\textbf{Blender:} \textit{i would encourage you to look up ` ` toxic relationships ' ' and try to make your relationships stronger .}\\
\textbf{GPT3:} \textit{I would encourage you to try to learn from your mistakes and meet some new people.}\vspace{1mm}\\

\textbf{\underline{Template+Retrieval-based (with generic prompting):}}\\

\textbf{Blender:} \textit{in addition to being open to meeting new friends , involve yourself in projects from which you ' ll feel satisfied by your own results .}\\
\textbf{GPT3:} \textit{I would encourage you to try to learn from your mistakes and meet some new people.}\vspace{1mm}\\

\textbf{\underline{Template+Retrieval-based (with N-gram prompting):}}\\
\textbf{Prompt:} \textit{it may be important to}\vspace{0.5mm}\\

\textbf{Blender:} \textit{it may be important to try to learn from your mistakes and meet a new person . : ) it may also be important for you to start engaging in things outside of the home that make you happy .}\\
\textbf{GPT3:} \textit{It may be important to try to learn from your mistakes and meet some new people}\vspace{1mm}\\

\midrule

\textbf{Advise without permission:} \textit{find something you enjoy and work towards it .
} \\

\textbf{Advise with permission (ground-truth):} \textit{It maybe helpful to find something you enjoy and work towards it .}\vspace{1mm}\\

\textbf{\underline{Template-based:}}\\

\textbf{Blender:} \textit{it would be a good idea to find something yourself and work towards it . . . i wonder if you have heard of this ?}\\
\textbf{GPT3:} \textit{It may be important to find something you enjoy and work towards it .}\vspace{1mm}\\

\textbf{\underline{Retrieval-based:}}\\

\textbf{Blender:} \textit{i think it would be valuable for you to start engaging in things outside of the home that make you happy .}\\
\textbf{GPT3:} \textit{Give yourself some credit for being awesome and tell yourself you can build new friendships .}\vspace{1mm}\\

\textbf{\underline{Template+Retrieval-based:}}\\

\textbf{Blender:} \textit{i think it would be valuable for you to start engaging in things outside of the home that make you happy .}\\
\textbf{GPT3:} \textit{It would be a good idea to find something you enjoy and work towards it .}\vspace{1mm}\\

\textbf{\underline{Template+Retrieval-based (with generic prompting):}}\\

\textbf{Blender:} \textit{i think it would be valuable for you to start engaging in things outside of the home that make you happy .}\\
\textbf{GPT3:} \textit{An option would be to find something you enjoy and work towards it .}\vspace{1mm}\\

\textbf{\underline{Template+Retrieval-based (with N-gram prompting):}}\\
\textbf{Prompt:} \textit{it maybe helpful to}\vspace{0.5mm}\\
\textbf{Blender:} \textit{it’s perhaps helpful to find something yourself and work towards it .}\\
\textbf{GPT3:} \textit{It maybe helpful to find something you enjoy and work towards it .}\vspace{1mm}\\

\midrule

\textbf{Advise without permission:} \textit{Talk to a therapist .} \\

\textbf{Advise with permission (ground-truth):} \textit{I encourage you to meet with a therapist and ask them how they might be able to help you . }\vspace{1mm}\\

\textbf{\underline{Template-based:}}\\

\textbf{Blender:} \textit{it would be a good idea to talk to one of the therapists . i wonder if you can talk to an therapist .}\\
\textbf{GPT3:} \textit{I encourage you to talk to a therapist .}\vspace{1mm}\\

\textbf{\underline{Retrieval-based:}}\\

\textbf{Blender:} \textit{I encourage you to meet with a therapist and ask them how they might be able to help you .}\\
\textbf{GPT3:} \textit{I encourage you to meet with a therapist and ask them how they might be able to help you}\vspace{1mm}\\

\textbf{\underline{Template+Retrieval-based:}}\\

\textbf{Blender:} \textit{i encourage you to meet with a therapist and ask them how they might be able to help you .}\\
\textbf{GPT3:} \textit{I encourage you to meet with a therapist and ask them how they might be able to help}\vspace{1mm}\\

\textbf{\underline{Template+Retrieval-based (with generic prompting):}}\\

\textbf{Blender:} \textit{i encourage you to meet with a therapist and ask them how they might be able to help you .}\\
\textbf{GPT3:} \textit{I encourage you to meet with a therapist and ask them how they might be able to help}\vspace{1mm}\\

\textbf{\underline{Template+Retrieval-based (with N-gram prompting):}}\\
\textbf{Prompt:} \textit{able to help you}\vspace{0.5mm}\\
\textbf{Blender:} \textit{i encourage you to meet with a therapist . they ' ll help you find some strategies or tools to make it easier .}\\
\textbf{GPT3:} \textit{I encourage you to meet with a therapist and ask them how they might be able to help}\vspace{1mm}\\

\bottomrule
%\end{tabular}
\end{tabularx}
\caption{Examples of rephrased sentences by the different rephraser models we developed}
\label{tab:examples}
\end{table*}

\section{Automatic Evaluation}
\label{sec:autoeval}

We used the the NLTK package to compute the BLEU \footnote{\url{https://www.nltk.org/_modules/nltk/translate/bleu_score.html}}, METEOR \footnote{\url{https://www.nltk.org/_modules/nltk/translate/meteor_score.html}}, and chrf \footnote{\url{https://www.nltk.org/_modules/nltk/translate/chrf_score.html}} scores. The ROUGE score and the BERTscore were computed using the rouge \footnote{\url{https://pypi.org/project/rouge/}}  and bert\_score \footnote{\url{https://pypi.org/project/bert-score/}}  python libraries, respectively. The POS distance was calculated as mentioned in the work by Tian et al. \shortcite{pos} following the code released by the authors on github. \footnote{\url{https://github.com/YouzhiTian/Structured-Content-Preservation-for-Unsupervised-Text-Style-Transfer/blob/master/POS_distance.py}} For computing the Word Mover Distance (WMD), we used Gensim’s implementation of the WMD. \footnote{\url{https://radimrehurek.com/gensim/auto_examples/tutorials/run_wmd.html}} We used sentence embeddings generated using Sentence-BERT \cite{sbert} to compute the cosine similarity between the original and rephrased text. Among the models the authors have proposed, we used the \textit{roberta-base-nli-stsb-mean-tokens} model, fine-tuned on the NLI \cite{A_large_annotated_corpus_for_learning_natural_language_inference} and STS benchmark (STSb) \cite{SemEval-2017_Task1_Semantic_Textual_Similarity_Multilingual_and_Crosslingual_Focused_Evaluation} datasets to generate the embeddings. All the automatic evaluation scores are reported for a single run. 

% WMD, POS, Cos sim

%To do: If you used existing packages (e.g., for preprocessing, for normalization, or for evaluation), did you report the implementation, model, and parameter settings used (e.g., NLTK, Spacy, ROUGE, etc.)? 
%\footnote{\url{}}

\section{Human Evaluation}
\label{sec:humaneval}

%The demographics of the UpWork workers. 

\begin{figure}[ht!]
  \centering
  \fbox{\includegraphics[width=\linewidth]{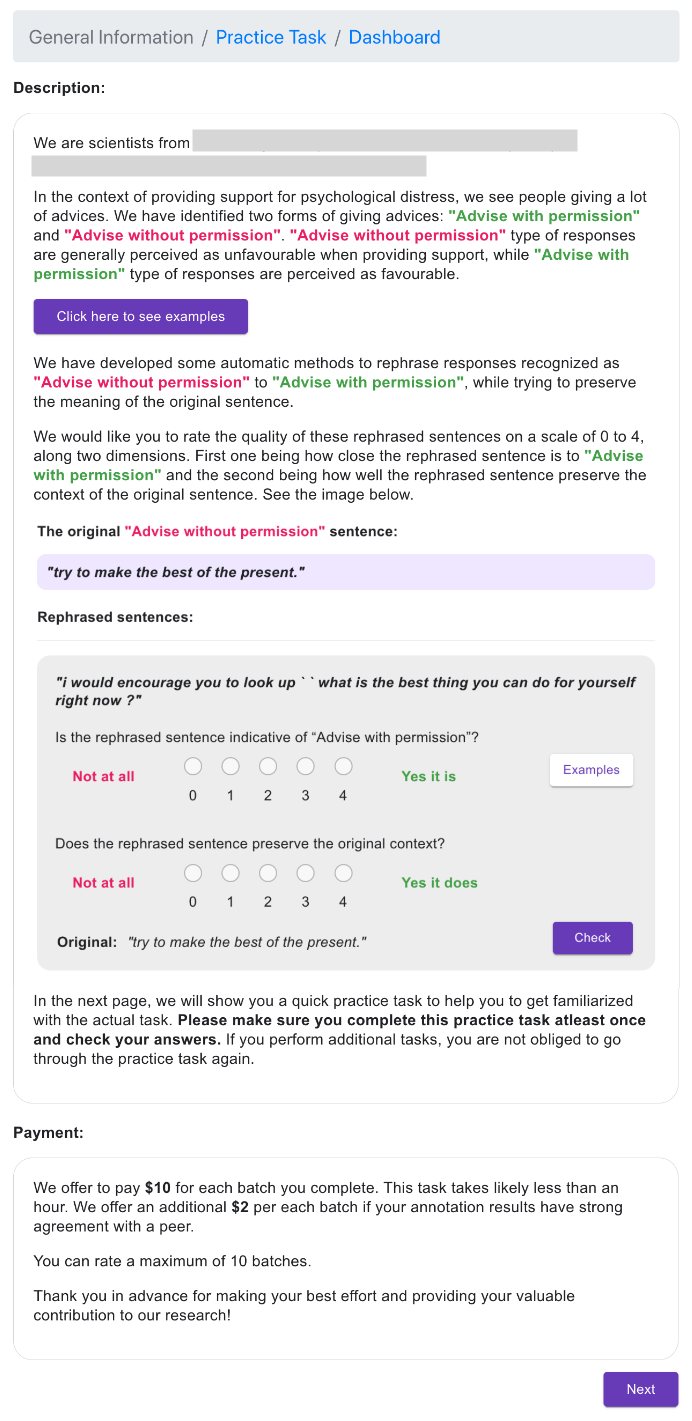}}
  \caption{Human evaluation task description.}
  \label{fig:description}
\end{figure}

\begin{figure}[ht!]
  \centering
  \fbox{\includegraphics[width=\linewidth]{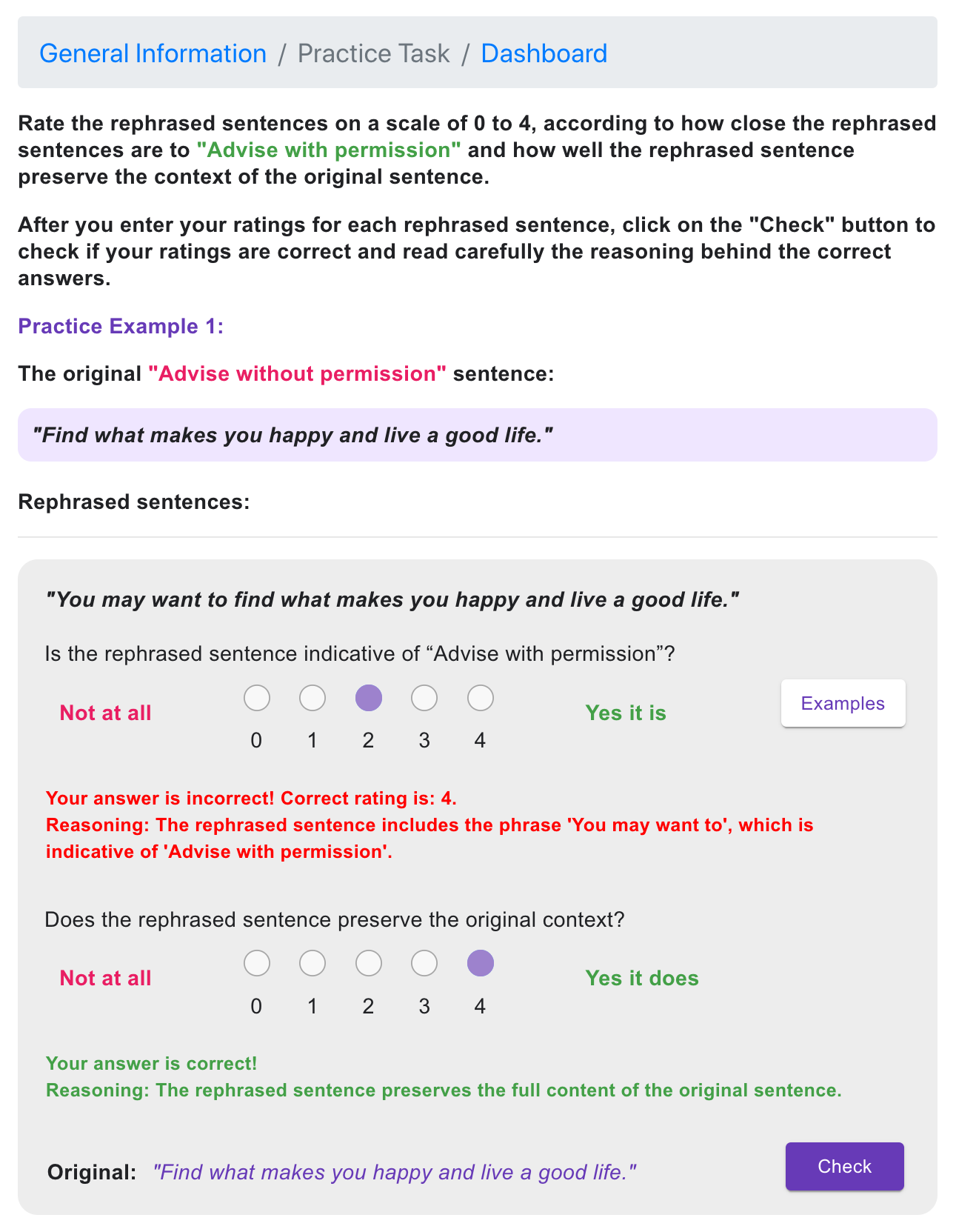}}
  \caption{Self-evaluating practice task offered to the counselors to get familiarized with the rating task.}
  \label{fig:practice}
\end{figure}

\begin{figure}[ht!]
  \centering
  \fbox{\includegraphics[width=\linewidth]{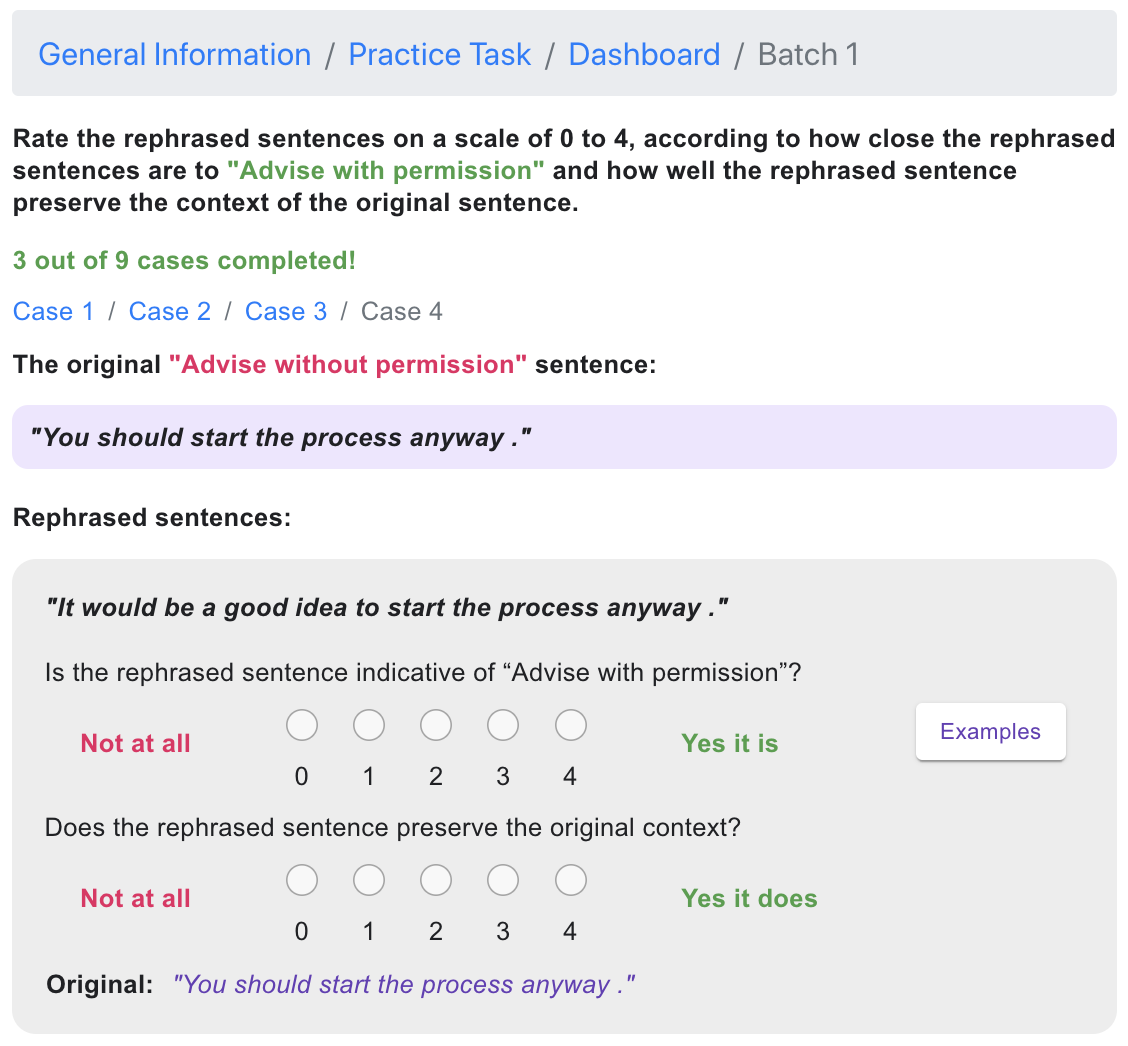}}
  \caption{The human evaluation task interface.}
  \label{fig:human-eval}
\end{figure}

Figures \ref{fig:description}, \ref{fig:practice}, and \ref{fig:human-eval} shows the user interfaces developed for the human evaluation task. The first one shows the task description, the second one shows the self-evaluating practice task designed to get the counselors familiarized with the rating task, and the last one shows the actual human evaluation task itself. 

%Figure \ref{fig:human-eval} shows the human evaluation task interface  that includes the question format used when the counselors were asked to rate the rephrased sentences.

% asked to rate of the rephrased sentences on a Likert scale ranging from 0 to 4 on each of the above dimensions. 

%Table \ref{tab:demographics} shows the demographic and other details of the counselors we recruited from UpWork for the rating task. 

%\begin{table}[ht!]
%\small
%\centering
%\begin{tabularx}{\linewidth}{X p{1cm} p{1cm} p{1cm}}
%\begin{tabularx}{\linewidth}{X}
%\begin{tabular}{l}
%\toprule

%\textbf{Counselor \#1}\vspace{1mm}\\

%Country of origin: Egypt\\
%Gender: Male\\

%\bottomrule
%\end{tabular}
%\end{tabularx}
%\caption{Demographic and other details of the counselors who participated in the rating task.}
%\label{tab:demographics}
%\end{table}

\section{Other Remarks}
\label{sec:remarks}

In human evaluation results, we observed in 97.5\% of the cases, the average scores obtained for style transfer strength are better than the average scores obtained for semantic similarity. This observation is invariant of the type of backbone model used in training. This implies template-based and retrieval-based methods used in creating pseudo parallel data to train the rephrasers make it easier for the rephrasers to generate rephrased sentences that reflect a particular style (in this case, \textit{Advise with permission}) than preserving the semantic meaning of the original sentence. This is a matter to be further investigated. To improve the scores on semantic similarity, future work can explore ways to take into account the context that precedes the sentence to be rephrased. In this way, though the rephrased version may not reflect exactly what was in the original sentence, it might still be able to generate rephrasings relevant to the preceding context. 

% It was observed from the human evaluation results, that compared to style transfer strength, the ratings obtained by the rephrasers for context similarity were relatively lower. 

It should be noted that the application of this work is not limited to improving chatbot responses for distress consolation. This could also be applied for the development of intelligent writing assistants that can suggest better responses when peers untrained in the practice of counseling attempt to respond to distress-related posts on peer support platforms such as Reddit.

\section{Distribution and Use of Artifacts}
\label{sec:artifacts}

The artifacts produced, including the datasets and the models, will be released under the CC BY-NC-SA 3.0 license \url{https://creativecommons.org/licenses/by-nc-sa/3.0}, providing only non-commercial access to the users. We use artifacts such as the CounselChat dataset, and pre-trained language architectures such as BERT \cite{bert}, RoBERTA \cite{roberta}, Blender \cite{blender}, and GPT3 \cite{gpt3} for research purposes only, which does not violate their intended use.  

\end{document}